\documentclass[11pt,letterpaper]{article}

\usepackage[hidelinks]{hyperref}
\usepackage{color,array}

\usepackage{graphicx}

\usepackage[margin=1.0in]{geometry}

\usepackage[utf8]{inputenc} 
\usepackage[T1]{fontenc}    
\usepackage{url}            
\usepackage{booktabs}       
\usepackage{amsfonts}       
\usepackage{nicefrac}       
\usepackage{microtype}      
\usepackage{xcolor}         

\usepackage{graphics} 
\usepackage{amsmath} 
\usepackage{amssymb}  
\usepackage{color,xspace}
\usepackage{graphicx}
\usepackage{subcaption}
\usepackage{cleveref}
\usepackage{bbm}
\usepackage{tikz}
\usepackage{epstopdf} 
\usepackage{ntheorem}
\usepackage{mdframed}   

\usepackage[T1]{fontenc}
\usepackage{lmodern}

\usepackage[ruled,vlined]{algorithm2e}
\usepackage{algorithmic}
\usepackage{booktabs}
\usepackage{tabularx}
\usepackage{multicol}
\usepackage{wrapfig}

\newtheorem{remark}{Remark}
\newtheorem{example}{Example}

\newcommand{\bx}{\mathbf{x}}
\newcommand{\by}{\mathbf{y}}

\newcommand{\bw}{\mathbf{w}}

\newcommand{\bz}{\mathbf{z}}
\newcommand{\blambda}{\boldsymbol{\lambda}}
\newcommand{\bmu}{\boldsymbol{\mu}}
\newcommand{\bnu}{\boldsymbol{\nu}}
\newcommand{\bxi}{\boldsymbol{\xi}}

\newcommand{\relu}{\mathrm{ReLU}}
\newcommand{\relaxed}{\mathrm{relaxed}}
\newcommand{\proj}{\mathrm{Proj}}

\DeclareMathOperator*{\argmax}{arg\,max}
\DeclareMathOperator*{\argmin}{arg\,min}

\title{{\LARGE \bf DeepSplit: Scalable Verification of Deep Neural Networks via Operator Splitting}}

\author{Shaoru Chen$^\dagger$, Eric Wong$^\dagger$, J. Zico Kolter, Mahyar Fazlyab 
\thanks{
Shaoru Chen is with the Department of Electrical and Systems Engineering, University of Pennsylvania, email: srchen@seas.upenn.edu; Eric Wong is with the Computer Science and Artificial Intelligence Laboratory, Massachusetts Institute of Technology, email: wongeric@mit.edu; 
J. Zico Kolter is with the Computer Science Department, Carnegie Mellon University, email: zkolter@cs.cmu.edu; 
Mahyar Fazlyab is with the Mathematical Institute for Data Science, Johns Hopkins University, email: mahyarfazlyab@jhu.edu.  \newline
$^\dagger$ The first two authors contribute equally to this paper. \newline
Codes of the presented method are available at~\url{https://github.com/ShaoruChen/DeepSplit}.
} 
}

\date{}

\begin{document}
\pagestyle{plain}
\maketitle

\begin{abstract}
Analyzing the worst-case performance of deep neural networks against input perturbations amounts to solving a large-scale non-convex optimization problem, for which several past works have proposed convex relaxations as a promising alternative. However, even for reasonably-sized neural networks, these relaxations are not tractable, and so must be replaced by even weaker relaxations in practice. In this work, we propose a novel operator splitting method that can directly solve a convex relaxation of the problem to high accuracy, by splitting it into smaller sub-problems that often have analytical solutions. The method is modular, scales to very large problem instances, and compromises of operations that are amenable to fast parallelization with GPU acceleration. {We demonstrate our method in bounding the worst-case performance of large convolutional networks in image classification and reinforcement learning settings, and in reachability analysis of neural network dynamical systems.
}
\end{abstract}

\section{Introduction}

Despite their superior performance, neural networks lack formal guarantees, raising serious concerns about their adoption in  safety-critical applications such as autonomous vehicles \cite{cao2019adversarial} and medical machine learning \cite{finlayson2019adversarial}. Motivated by this drawback, 
%
there has been an increasing interest in developing tools to verify desirable properties for neural networks, such as robustness to adversarial attacks.

%

Neural network verification refers to the problem of verifying whether the output of a neural network satisfies certain properties for a bounded set of input perturbations. This problem can be framed as optimization problems of the form
\begin{align} \label{eq: maximization problem}
J^\star \leftarrow \mathrm{minimize} \quad J(f(x)) \quad \text{subject to}  \quad x \in \mathcal{X},
\end{align}
where $f$ is given by a deep neural network, $J$ is a real-valued function representing a performance measure (or a specification), and $\mathcal{X}$ is a set of inputs to be verified. In this formulation, verifying the neural network amounts to certifying whether the optimal   value of \eqref{eq: maximization problem} is bounded below by a certain threshold. 

{As an example, consider a classification problem with $n_f$ classes, in which for a data point $x \in \mathbb{R}^{n_x}$, $f(x) \in \mathbb{R}^{n_f}$ denotes the vector of scores for all the classes. The classification rule is $C(x) = \argmax_{1 \leq i \leq n_f} e_i^\top f(x)$ where $e_i$ denotes the $i$-th standard basis. Given a correctly classified $x^\star \in \mathbb{R}^{n_x}$ and a perturbation set $\mathcal{X} \subset \mathbb{R}^{n_x}$ that contains $x$, we say that $f$ is locally robust at $x^\star$ with respect to $\mathcal{X}$ if $C(x)=C(x^\star)$ for all $x \in \mathcal{X}$. Verifying the local robustness at $x^\star$ then amounts to verifying that the optimal values of the following $n_f-1$ optimization problems
\begin{align} \label{eq:class_verify}
    \underset{x \in \mathcal{X}}{\operatorname{minimize}} &\quad (e_{i^\star} - e_i)^\top f(x), \quad i \neq i^\star,
\end{align}
are positive, where $i^\star=C(x^\star)$ is the class index of $x^\star$. Problem \eqref{eq:class_verify} is an instance of \eqref{eq: maximization problem} where $J$ is a linear function.} 

Problem \eqref{eq: maximization problem} is large-scale and non-convex, making it extremely difficult to solve efficiently--both in terms of time and memory. 
%
%
For $\relu$ activation functions and linear objectives, the problem in \eqref{eq: maximization problem} can be cast as a Mixed-Integer Linear Program (MILP) \cite{lomuscio2017approach,cheng2017maximum,dutta2018output,fischetti2018deep}, which can be solved for the global solution via, for example, Branch-and-Bound (BaB) methods. While  we  do  not  expect  these  approaches  to  scale  to  large problems,  for  small  neural  networks they can still be practical.

Instead of solving \eqref{eq: maximization problem} for its global minimum, one can instead find \emph{guaranteed lower bounds} on the optimal value via convex relaxations, such as Linear Programming (LP) \cite{kolter2017provable} and Semidefinite Programming (SDP) \cite{raghunathan2018semidefinite,fazlyab2019efficient,fazlyab2020safety}. 
Verification methods based on convex relaxations are sound but incomplete, i.e., they are guaranteed to detect all {false negatives} but also produce false positives, whose rate depends on the tightness of the relaxation. 
Although convex relaxations are polynomial-time solvable (in terms of number of decision variables), in practice they are not computationally tractable for large-scale neural networks. To improve scalability, these relaxations must typically be further relaxed \cite{kolter2017provable,weng2018towards,zhang2018efficient,dvijotham2018dual,bunel2020lagrangian}.

\paragraph{Contributions} 
In this work, we propose an algorithm to solve LP relaxations of \eqref{eq: maximization problem} for their global solution and for large-scale feed-forward neural networks.
%
%
%
Our starting point is to express \eqref{eq: maximization problem} as a constrained optimization problem whose constraints are imposed by the forward passes in the network. We then introduce additional decision variables and consensus constraints that naturally split the corresponding problem into independent subproblems, which often have closed-form solutions. 
Finally, we employ an operator splitting technique based on the Alternating Direction Method of Multipliers (ADMM)~\cite{boyd2011distributed}, to solve the corresponding Lagrangian relaxation of the problem. 
This approach has several favorable properties. First, the method requires minimal parameter tuning and relies on simple operations, which scale to very large problems and can achieve a good trade-off between runtime and solution accuracy. Second, all the solver operations are amenable to fast parallelization with GPU acceleration. Third, our method is fully modular and applies to standard network architectures. 

We employ our method to compute exact solutions to LP relaxations on the worst-case performance of adversarially trained deep networks, with a focus on networks whose convex relaxations {are difficult} to solve due to their size. Specifically, we perform extensive experiments in the $\ell_\infty$ perturbation setting, where we verify robustness properties of image classifiers for CIFAR10 and deep Q-networks (DQNs) in Atari games \cite{zhang2020robust}. Our method is able to solve LP relaxations at scales that are too large for exact {MILP} verifiers, SDP relaxations, or commercial LP solvers such as Gurobi. {We also demonstrate the use of our method in  computing reachable set over-approximations of a neural network dynamical system over a long horizon, where our method is compared with the state-of-the-art BaB method~\cite{wang2021beta}.}

\subsection{Related work}

\paragraph{Convex relaxations} 
LP relaxations are relatively the most scalable form of convex relaxations \cite{ehlers2017formal}. However, even solving LPs can become computationally prohibitive for small convolutional networks \cite{salman2019convex}.
%
%
One line of work studies computationally cheaper but looser bounds of the LP relaxation \cite{kolter2017provable}, which {we denote as linear bounds}, and have been extended to larger and more general networks and settings \cite{wong2018scaling, weng2018towards, zhang2018efficient, xu2020automatic}. These bounds tend to be loose unless optimized during training, which typically comes at a significant cost to standard performance. Further work has aimed to tighten these bounds \cite{singh2019beyond,tjandraatmadja2020convex, de2021scaling}, however these works focus primarily on small convolutional networks and struggle to scale to more typical deep networks. Other work has studied the limits of these convex relaxations on these small networks using vast amounts of CPU-compute \cite{salman2019convex}. Recent SDP-based approaches \cite{dathathri2020enabling} can produce much tighter bounds on these small networks. 

\paragraph{Lagrangian-based bounds}
Related to our work is that which solves the Lagrangian of the LP relaxation \cite{dvijotham2018dual,bunel2020lagrangian}, which can tighten the bound but do not aim to solve the LP exactly due to relatively slow convergence.
However, these works primarily study small networks whose LP relaxation can still be solved exactly with Gurobi. Although these works could in theory be used on larger networks, only the faster, linear bounds-based methods \cite{xu2020automatic} have demonstrated applicability to standard deep learning architectures. In our work, we solve the LP relaxation exactly in \emph{large} network settings that previously have only been studied with loose bounds of the LP relaxation such as LiRPA \cite{xu2020automatic}. 

\paragraph{Operator splitting methods} 
Operator splitting, and in particular the ADMM method, is a powerful technique in solving structured convex optimization problems and has applications in numerous settings ranging from optimal control \cite{o2013splitting} to training neural networks \cite{taylor2016training}. These methods scale well with the problem dimensions, can exploit sparsity in the problem data efficiently \cite{zheng2017fast}, are amenable to parallelization on GPU \cite{schubiger2020gpu}, and have well-understood convergence properties under minimal regularity assumptions \cite{boyd2011distributed}. The benefit of ADMM as an alternative to interior-point solvers has been shown in various classes of optimization problems \cite{o2016conic}. Our operator splitting method is specifically tailored for neural network verification in order to fully exploit the problem structure.

\paragraph{Complete verification methods} 
These methods verify properties of deep networks exactly (i.e., they find the optimal value $J^\star$ and a global solution $x^\star$) using methods such as Satisfiability Modulo Theories (SMT) solvers \cite{scheibler2015towards,ehlers2017formal,katz2017reluplex} and MILP solvers \cite{lomuscio2017approach, cheng2017maximum, dutta2018output, fischetti2018deep}. Complete verification methods typically rely on BaB algorithms \cite{bunel2017unified}, in which the verification problem is divided into subproblems (branching) that can be verified using incomplete verification methods (bounding) \cite{bunel2020branch}. However, these methods have a worst-case exponential runtime and have difficulty scaling beyond relatively small convolutional networks. Motivated by this, several recent works have been reported on improving the scalability of BaB methods by developing custom solvers in the bounding part to compute cheaper intermediate bounds and hence, speed up their practical running time \cite{bunel2020lagrangian,de2021improved, xu2021fast, wang2021beta}. {In the same spirit, our proposed method for solving the large-scale LP relaxations can be potentially used as a bounding subroutine in BaB for complete verification which we leave for future research. }


\paragraph{Notation} We denote the set of real numbers by $\mathbb{R}$, the set of nonnegative real numbers by $\mathbb{R}_{+}$, the set of real $n$-dimensional vectors by $\mathbb{R}^n$, the set of $m\times n$-dimensional real-valued matrices by $\mathbb{R}^{m\times n}$, 
and the $n$-dimensional identity matrix by $I_n$. 
The $p$-norm ($p \geq 1$) is denoted by $\|\cdot\|_p \colon \mathbb{R}^n \to \mathbb{R}_{+}$. 
%
%
For a set $\mathcal{S}$, we define the indicator function $\mathbb{I}_{\mathcal{S}}(x)$ of $\mathcal{S}$ as $\mathbb{I}_{\mathcal{S}}(x)=0$ if $x \in \mathcal{S}$ and $\mathbb{I}_{\mathcal{S}}(x)= +\infty$ otherwise. Given a function  $f \colon \mathcal{X} \to \mathcal{Y}$, the graph of $f$ is the set $\mathcal{G}_{f} = \{(x,f(x)) \mid x \in \mathcal{X}\}$.

\section{Neural network verification via operator splitting} \label{sec::Scalable Neural Network Verification via Operator Splitting}
We consider an $\ell$-layer feed-forward neural network $f(x) \colon \mathbb{R}^{n_0} \to \mathbb{R}^{n_\ell}$ described by the following recursive equations,
\begin{align}  \label{eq: DNN model}
    \begin{aligned}
    x_0 &= x,\\  x_{k+1} &= \phi_k(x_k), \quad  k=0,\cdots,\ell-1,\\
    f(x) &= x_{\ell}
    \end{aligned}
\end{align}
where $x_0 \in \mathbb{R}^{n_0}$  is the input to the neural network, $x_{k} \in \mathbb{R}^{n_k}$ is the input to the $k$-th layer, and $\phi_k \colon \mathbb{R}^{n_k} \to \mathbb{R}^{n_{k+1}}$ is the operator of the $k$-th layer, which can represent any commonly-used operator in deep networks, such as linear (convolutional) layers~\footnote{Convolution is a linear operator and conceptually it can be analyzed in the same way as for linear layers.}, MaxPooling units, and activation functions.
%

Given the neural network $f$, a specification function $J \colon \mathbb{R}^{n_\ell} \mapsto \mathbb{R}$, and an input set $\mathcal{X} \subset \mathbb{R}^{n_0}$, we say that $f$ satisfies the specification $J$ if $J(f(x)) \geq 0$ for all $x \in \mathcal{X}$. This is equivalent to verifying that the optimal value of \eqref{eq: maximization problem} is non-negative. We assume $\mathcal{X} \subset \mathbb{R}^{n_0}$ is a closed convex set and $J \colon \mathbb{R}^{n_{\ell}} \to \mathbb{R} \cup \{+\infty\}$ is a convex function.
%

%



Using the sequential representation of the neural network in \eqref{eq: DNN model}, we may rewrite the optimization problem in \eqref{eq: maximization problem} as the following constrained optimization problem,
\begin{equation}\label{eq: nn verification problem constrained}
\begin{aligned} 
J^\star \leftarrow \text{minimize} &\quad J(x_{\ell}) \\ 
\text{subject to } &\quad x_{k+1}=\phi_k(x_{k}), \  k=0,\cdots,\ell-1, \\ 
&\quad x_0 \in \mathcal{X},
\end{aligned}
\end{equation}
%
with $n:=\sum_{k=0}^{\ell} n_k$ decision variables $x_0,\cdots,x_{\ell}$.
We can rewrite \eqref{eq: nn verification problem constrained} equivalently as
\begin{equation}\label{eq: nn verification problem constrained graph form}
\begin{aligned} 
J^\star \leftarrow \text{minimize}& \quad J(x_{\ell})  \\ \text{subject to} &\quad (x_{k},x_{k+1})\in \mathcal{G}_{\phi_k}, \ k=0,\cdots,\ell-1,  \\
& \quad x_0 \in \mathcal{X},
\end{aligned}
\end{equation}
where 
$$\mathcal{G}_{\phi_k} = \{(x_k,x_{k+1}) \mid x_{k+1} = \phi_k(x_k), \ \underline{x}_k \leq x_k \leq \bar{x}_k\},$$ is the graph of $\phi_k$. Here $\underline{x}_k$ and $\bar{x}_k$ are \emph{a priori} known bounds on $x_k$ when $x_0 \in \mathcal{X}$~\footnote{See Section~\ref{sec:activation_layers} for comments on finding the bounds $\underline{x}_k$ and $\bar{x}_k$.}. 

The problem in \eqref{eq: nn verification problem constrained graph form} is non-convex due to presence of nonlinear operators in the network, such as activation layers, which render the set $\mathcal{G}_{\phi_k}$ nonconvex. By over-approximating $\mathcal{G}_{\phi_k}$ by a convex set (or ideally by its convex hull), we arrive at a direct layer-wise convex relaxation of the problem, {in which each two consecutive variables $(x_k,x_{k+1})$ are sequentially coupled together. Solving this relaxation directly cannot exploit this structure and is hence unable to scale to even medium-sized neural networks \cite{salman2019convex}. In this section, by exploiting the sequential structure of the constraints, and introducing auxiliary decision variables (variable splitting), we propose a reformulation of \eqref{eq: nn verification problem constrained} whose convex relaxation can be decomposed into smaller sub-problems that can be solved efficiently and in a scalable manner.}

\subsection{Variable splitting}
\begin{figure}
    \centering
    \includegraphics[width=0.5\textwidth]{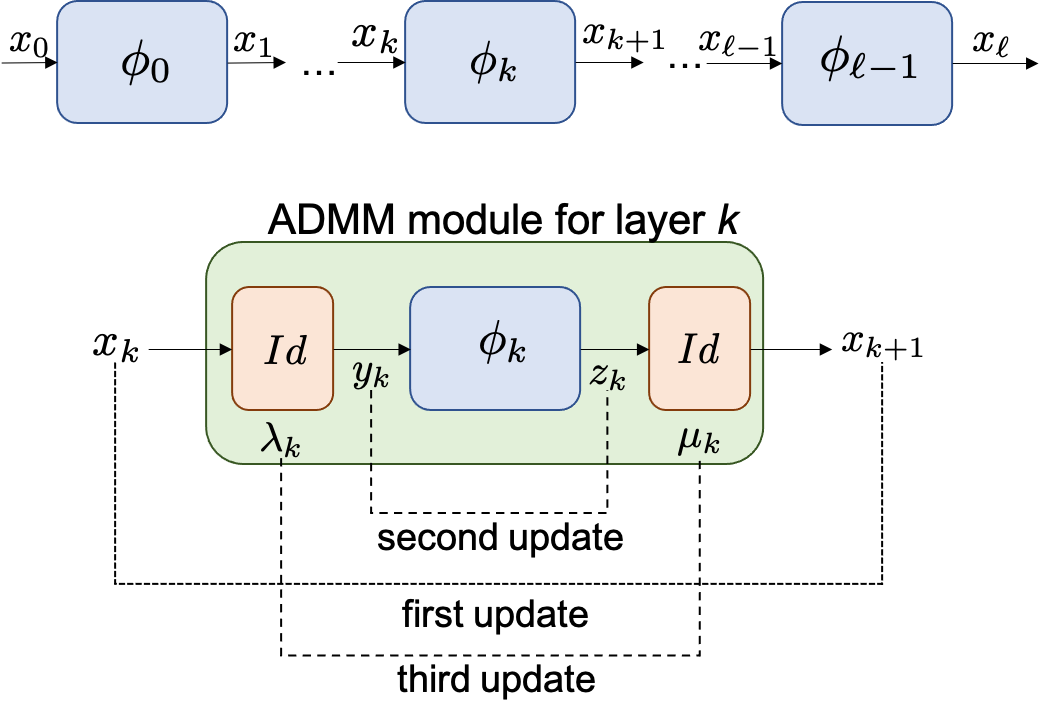}
    \caption{Illustration of the network structure (top) and DeepSplit computation module for a generic layer (bottom). Adding identity layers in between the neural network layers decouples the variables $x_k$ and allows processing them independently.}
    \label{fig:admm_module}
\end{figure}

By introducing the intermediate variables $y_k$ and $z_k$, we can rewrite \eqref{eq: nn verification problem constrained} as
%
\begin{alignat}{3} \label{eq: verification problem split}
J^\star \leftarrow &\text{minimize} \quad && J(x_\ell)  \\ \notag
&\text{subject to } \quad && y_{k} = x_{k},  \quad && k=0,\cdots,\ell-1 \\ \notag
& \quad && z_k=\phi_k(y_k),  \quad &&k=0,\cdots,\ell-1 \\ \notag
& \quad && x_{k+1} = z_{k},  \quad &&k=0,\cdots,\ell-1 \\ \notag
& \quad && x_0 \in \mathcal{X},
\end{alignat}
which has now $3n-n_0-n_{\ell}$ decision variables. Intuitively, we have introduced additional ``identity layers'' between consecutive layers (see Figure \ref{fig:admm_module}). By overapproximating $\mathcal{G}_{\phi_k}$ by a convex set $\mathcal{S}_{\phi_k}$, we obtain the convex relaxation
\begin{alignat}{3} \label{eq: verification problem split convexified}
J_{\mathrm{relaxed}}^\star \leftarrow &\text{minimize} \quad && J(x_\ell)  \\ \notag
&\text{subject to } \quad && y_{k} = x_{k},  \quad && k=0,\cdots,\ell-1 \\ \notag
& \quad && (y_k,z_k) \in \mathcal{S}_{\phi_k},  \quad &&k=0,\cdots,\ell-1 \\ \notag
& \quad && x_{k+1} = z_{k},  \quad &&k=0,\cdots,\ell-1 \\ \notag
& \quad && x_0 \in \mathcal{X},
\end{alignat}
%
for which $J_{\mathrm{relaxed}}^\star \leq J^\star$. This form is known as consensus as $y_k$ and $z_{k-1}$ are just copies of the variable $x_k$. As shown below, this ``overparameterization'' allows us to split the optimization problem into smaller sub-problems that can be solved in parallel and often in closed form.

\ifdefined\arxiv
  \def\figwidth{0.5\linewidth}
\else
  \def\figwidth{1.0\linewidth}
\fi

\begin{figure}
	\begin{subfigure}{0.45 \columnwidth}
		\includegraphics[width = \linewidth]{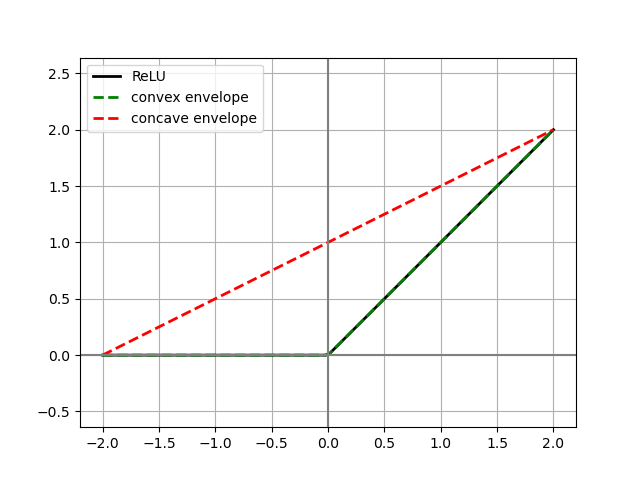}
	\end{subfigure}
	\hfill
	\begin{subfigure}{0.45 \columnwidth}
		\includegraphics[width = \linewidth]{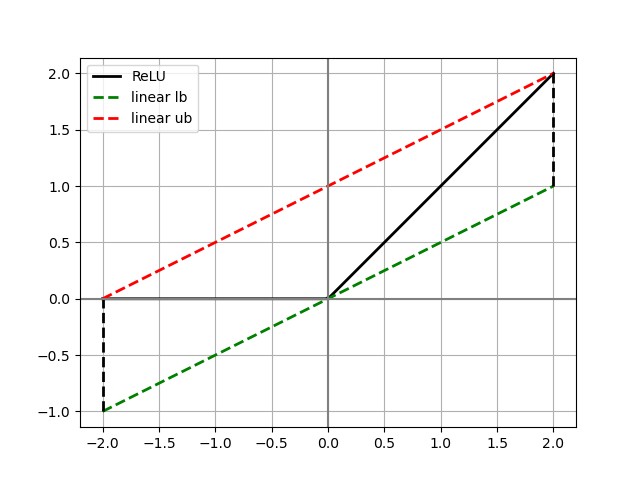}
	\end{subfigure}
	\caption{Over-approximation of the graph of ReLU function by convex hull (left) and linear bounds (right). }
	\label{fig:conv_over_approx}
\end{figure}

%
%
%
%
\subsection{Lagrangian relaxation and operator splitting}
We use $\bx=(x_0,\cdots,x_{\ell})$, $\by=(y_0,\cdots,y_{\ell-1})$ and $\bz=(z_0,\cdots,z_{\ell-1})$ to denote  the concatenated variables. By relaxing the equality constraints with Lagrangian multipliers, we define the augmented Lagrangian for \eqref{eq: verification problem split convexified} as follows,
\begin{equation} \label{eq: augmented lagrangian}
\begin{aligned}
\mathcal{L}(\bx,\by,\bz,\blambda,\bmu) &= J(x_{\ell}) \!+\! \sum_{k=0}^{\ell-1} \mathbb{I}_{\mathcal{S}_{\phi_k}}(y_{k}, z_{k})\!+\!\mathbb{I}_{\mathcal{X}}(x_0) \\ 
&\quad + (\rho/2)\sum_{k=0}^{\ell-1} \left( \|x_k-y_k+\lambda_k\|_2^2-\|\lambda_k\|_2^2\right) + (\rho/2)\sum_{k=0}^{\ell-1} \left( \|x_{k+1}-z_k+\mu_k\|_2^2-\|\mu_k\|_2^2\right). \end{aligned}
\end{equation}
where $\blambda =(\lambda_0,\cdots,\lambda_{\ell-1})$ and $\bmu = (\mu_0,\cdots,\mu_{\ell-1})$ are the scaled dual variables (by $1/\rho$) and $\rho>0$ is the augmentation constant.
Note that we have only relaxed the equality constraints in~\eqref{eq: verification problem split convexified}, and the constraints describing the sets $\mathcal{S}_{\phi_k}$ as well as the input set $\mathcal{X}$ are kept intact. Furthermore, the inclusion of augmentation will render the dual function differentiable, and hence, easier to optimize.
%

For the Lagrangian in \eqref{eq: augmented lagrangian}, the dual function, which provides a lower bound to $J^\star_{\relaxed}$, is given by $ g(\blambda,\bmu) = \inf_{(\bx,\by,\bz)} \ \mathcal{L}(\bx,\by,\bz,\blambda,\bmu)$. The best lower bound can then be found by maximizing the dual function. \footnote{If strong duality holds, then this best lower bound would match $J^\star_{\relaxed}$. As shown in \cite{salman2019convex}, strong duality holds under mild conditions.}
However, solving the inner problem jointly over $(\bx,\by,\bz)$ to find the dual function is as difficult as solving a direct convex relaxation of \eqref{eq: nn verification problem constrained}. Instead, we split the primal variables $(\bx,\by,\bz)$ into $\bx$ and $(\by,\bz)$ and apply the classical ADMM algorithm to obtain the following iterations (shown in Figure~\ref{fig:admm_module}) for updating the primal and dual variables,
\begin{subequations} \label{eq: ADMM}
	\begin{align}
	\bx^{+} &\in  \mathrm{argmin}_{\bx}\  \mathcal{L}(\bx,\by,\bz,\blambda,\bmu) \label{eq: ADMM1} \\ 
	(\by^{+},\bz^{+}) &\in  \mathrm{argmin}_{(\by,\bz)} \ \mathcal{L}(\bx^{+},\by,\bz,\blambda,\bmu) \label{eq: ADMM2} \\ 
	(\blambda^{+},\bmu^{+})&\!=\! (\blambda,\bmu) \!+\! \nabla_{(\blambda,\bmu)} \ \mathcal{L}(\bx^{+},\by^{+},\bz^{+},\blambda,\bmu). \label{eq: ADMM3}
	\end{align}
\end{subequations}
As we show below, the Lagrangian has a separable structure by construction that can be exploited in order to efficiently implement each step of \eqref{eq: ADMM}.

\subsection{The $\bx$-update}
The Lagrangian in \eqref{eq: augmented lagrangian} is separable across the $x_k$ variables; hence, the minimization in \eqref{eq: ADMM1} can be done independently for each $x_k$. Specifically, for $k=0$, we obtain the following update rule for $x_0$,
\begin{subequations} \label{eq: ADMM x update breakdown}
\begin{align} \label{eq: admm first update 0}
    x_0^{+} = \proj_{\mathcal{X}}(y_0-\lambda_0).
\end{align}
Projections onto the $\ell_\infty$ and $\ell_2$ balls can be done in closed-form. For the $\ell_1$ ball, we can use the efficient projection scheme proposed in \cite{duchi2008efficient}, which has $\mathcal{O}(n_0)$ complexity in expectation. For subsequent layers, we obtain the updates
	\begin{align} \label{eq: admm first update}
	x_k^{+} &= \frac{1}{2}(y_k - \lambda_k + z_{k-1}-\mu_{k-1}), \ k = 1, \cdots, \ell -1, \\
	x_{\ell}^{+} \!&=\! \argmin_{x_{\ell}} \ J(x_{\ell}) \!+\! \frac{\rho}{2}\|x_{\ell}\!-\!z_{\ell-1}\!+\!\mu_{\ell-1}\|_2^2.
	\end{align}
\end{subequations}
For convex $J$ and $\rho>0$, the optimization problem for updating $x_{\ell}$ is strongly convex with a unique optimal solution. Indeed, its solution is the proximal operator of $J/\rho$ evaluated at $z_{\ell-1}-\mu_{\ell-1}$. For the special case of linear objectives, $J(x_\ell) = c^\top x_\ell$, we obtain the closed-form solution 
\begin{align*}
    x_\ell^{+} = -\frac{1}{\rho} c + (z_{\ell-1} - \mu_{\ell-1}).
\end{align*}

\subsection{The $(\by,\bz)$-update}
\label{sec:yz_update}
Similarly, the Lagrangian is also separable across the $(y_k,z_k)$ variables. Updating these variables in \eqref{eq: ADMM2} corresponds to the following projection operations per layer,
\begin{align} \label{eq:yz_proj}
(y_{k}^{+}, z_{k}^{+}) &= \proj_{\mathcal{S}_{\phi_k}}(x_k^{+} + \lambda_k, x_{k+1}^{+}+\mu_k),
\end{align}
for $k=0,\cdots,\ell-1$. Depending on the type of the layer (linear, activation, convolution, etc.), we obtain different projections which we describe below.
\subsubsection{Affine layers}
\label{sec:affine_layers}
Suppose $\phi_k(y_k) = W_k y_k + b_k$ is an affine layer representing a fully-connected, convolutional, or an average pooling layer. Then the graph of $\phi_k$ is already a convex set given by
%
    $\mathcal{G}_{\phi_k} = \{(y_k,z_k) \mid z_k = W_k y_k + b_k\}$. 
%
%
%
Choosing $\mathcal{S}_{\phi_k} = \mathcal{G}_{\phi_k}$, the projection in \eqref{eq:yz_proj} takes the form
\begin{equation} \label{eq:affine_proj}
\begin{aligned}
y_k^{+} &= (I_{n_{k}} \!+\! W_k^\top W_k)^{-1}(x_k^{+}+\lambda_k \!+\! W_k^\top (x_{k+1}^{+}\!+\!\mu_k\!-\!b_k)), \\
z_k^{+} &= W_k y_k^{+} + b_k.
\end{aligned}
\end{equation}
The matrix $(I_{n_k}+W_k^\top W_k)^{-1}$ can be pre-computed and cached for subsequent iterations. We can do this efficiently for convolutional layers using the fast Fourier transform (FFT) which we discuss in Appendix~\ref{app:convolutions} and~\ref{app:FFT}.

\subsubsection{Activation layers}
\label{sec:activation_layers}
For an activation layer of the form $\phi(x) := [\varphi_1(x_1) \ \cdots \ \varphi_{n}(x_n)]^\top$, the convex relaxation of $\mathcal{G}_{\phi}$ is given by the Cartesian product of individual convex relaxations i.e.,  $\mathcal{S}_\phi = \mathcal{S}_{\varphi_1} \times \cdots \times \mathcal{S}_{\varphi_n}$.
For a generic activation function $\varphi \colon \mathbb{R} \to \mathbb{R}$, suppose we have a concave upper bound $\bar{\varphi}$ and a convex lower bound $\underline{\varphi}$ on $\varphi$ over an interval $I = [\underline{x},\bar{x}]$, i.e., 
$
\underline{\varphi}(x) \leq \varphi(x) \leq \bar{\varphi}(x) \ \forall \ x \in [\underline{x},\bar{x}]$. A convex overapproximation of $\mathcal{G}_{\varphi}$ is 
\begin{align}
\mathcal{S}_{\varphi} \! = \! \{(x,y) \mid \underline{\varphi}(x) \leq y \leq \bar{\varphi}(x), \ \underline{x} \leq x \leq \bar{x} \},
\end{align}
which turns out to be the convex hull of  $\mathcal{G}(\varphi)$ when $\bar{\varphi}$ and $\underline{\varphi}$ are concave and convex envelopes of $\varphi$, respectively. The assumed pre-activation bounds $\underline{x}$ and $\bar{x}$ used to relax the activation functions can be obtained a priori via a variety of existing techniques such as linear bounds~\cite{kolter2017provable,zhang2018efficient,zhang2020towards} which propagate linear lower and upper bounds on each activation function (see Figure~\ref{fig:conv_over_approx}) throughout the network in a fast manner. 

\begin{figure}[htb]
	\centering
	\includegraphics[width = 0.5\columnwidth]{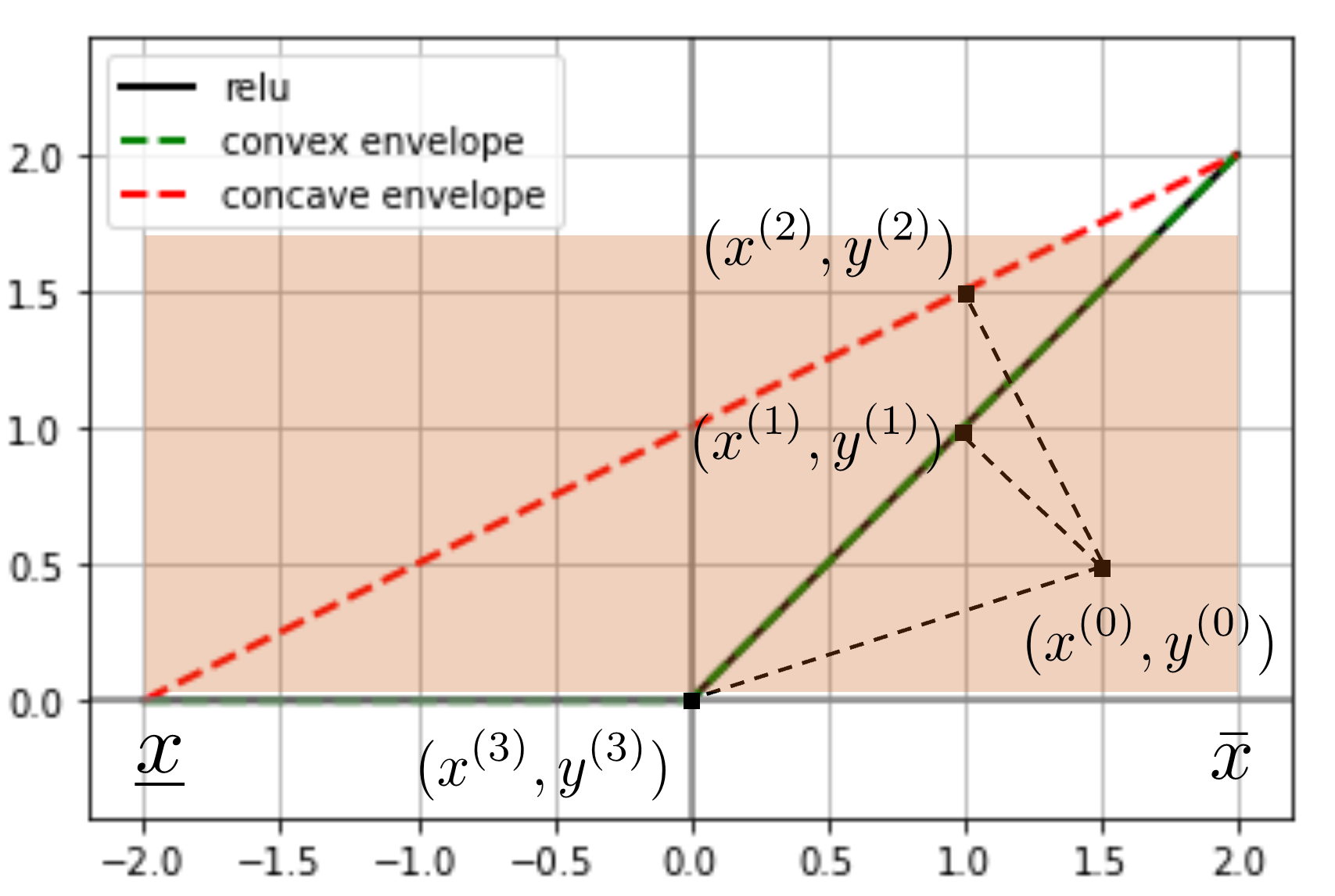}
	\caption{Projection of a point $(x_0,y_0)$ onto the convex hull of the ReLU function $y=\max(0,x)$ over the interval $[\underline{x},\bar{x}]$. We first project $(x_0,y_0)$ onto all facets of the convex hull and then select the point with minimal distance to $(x_0,y_0)$.}
	\label{fig:relu_proj}
\end{figure}

\begin{example}[ReLU activation function.]
\label{example:relu}
\normalfont Consider the $\relu$ activation function $\varphi(x)=\max(x,0)$ over the interval $[\underline{x},\bar{x}]$. When $\underline{x}<0<\bar{x}$, the $\relu$ function admits the envelopes $\underline \varphi(x) = \max(0,x)$, $\bar \varphi(x)= \underline{y} + \frac{\bar{y}-\underline{y}}{\bar{x}-\underline{x}}(x-\underline{x})$ on $[\underline{x}, \bar{x}]$, where $\underline{y}=\max(0,\underline{x})$ and $\bar{y}=\max(0,\bar{x})$ \cite{ehlers2017formal,kolter2017provable}. In this case, the projection of a point $(x^{(0)},y^{(0)})$ onto the convex hull of $G_\varphi$, which is a triangle shown in Figure~\ref{fig:conv_over_approx}, has a closed-form solution.
%
%
Letting $s = \frac{\bar{y}-\underline{y}}{\bar{x}-\underline{x}}$, we first project $(x^{(0)},y^{(0)})$ onto each facet of the triangle and then select the point with the minimal distance:
\begin{align*}
\begin{aligned}
x^{(1)} &= \min(\max(\frac{x^{(0)}+y^{(0)}}{2},0),\bar{x}), \quad y^{(1)}= x^{(1)}, \\
x^{(2)} &= \min(\max(\frac{x^{(0)}+sy^{(0)}+s(s\underline{x}-\underline{y})}{s^2+1},\underline{x}),\bar{x}), \\
y^{(2)}&= \frac{s(x^{(0)}-\underline{x})+s^2y^{(0)}+\underline{y}}{s^2+1},\\
x^{(3)} &= \min(\max(0,x^{(0)}),\underline{x}), \quad y^{(3)}=0.
\end{aligned}
\end{align*}
The projected point is $(x',y') = (x^{(i^\star)},y^{(i^\star)})$, where $i^\star = \arg\min_{1 \leq i \leq 3} \allowdisplaybreaks \sqrt{(x^{(0)}-x^{(i)})^2+(y^{(0)}-y^{(i)})^2}$. See Figure~\ref{fig:relu_proj} for illustration. 

When either $\underline{x} \geq 0$ or $\bar{x} \leq 0$, the graph $G_\varphi$ of the ReLU function becomes a line segment which allows closed-form projection of a point. 
\end{example} 

\subsection{The $(\blambda,\bmu)$-update}
Finally, we update the scaled dual variables as follows,
\begin{alignat}{2}
\begin{aligned} \label{eq:dual_update}
\lambda_{k}^{+} &= \lambda_{k} + (x_k^+ - y_k^+), \ &&k=0,\cdots,\ell-1,\\
\mu_k^{+} &= \mu_k +(x_{k+1}^{+}-z_k^{+}), \ && k=0,\cdots, \ell-1.
\end{aligned}
\end{alignat}
The DeepSplit Algorithm is summarized in Algorithm~\ref{DeepSplit Algorithm}. 

\begin{algorithm}[htb]
\SetAlgoLined
\KwData{neural network $f$ (Eq. \eqref{eq: DNN model}), bounded convex input set $\mathcal{X}$, convex function $J$.}
\KwResult{lower bound $J^\star_{\mathrm{relaxed}}$ on Problem \eqref{eq: maximization problem}.}
 \textbf{Initialization:}{ $x_0 \in \mathcal{X}$, $x_{k+1}=\phi_k(x_k)$, $y_k = x_k, \ z_k=x_{k+1}$, $\lambda_k = 0, \mu_k = 0$, $k=0,\cdots,\ell-1$, augmentation constant $\rho>0$.} \\
 \Repeat{stopping criterion is met (see Section~\ref{sec:convergence})}{
{\textbf{Step 1:} $\bx$-update \eqref{eq: ADMM x update breakdown}} \\
{ \textbf{Step 2:} $(\by,\bz)$-update \eqref{eq:yz_proj} }\\
 {\textbf{Step 3:} dual update \eqref{eq:dual_update} }
 }
 \textbf{Output:} {$J(x_\ell)$}
 \caption{DeepSplit Algorithm}
 \label{DeepSplit Algorithm}
\end{algorithm}

\subsection{Convergence and stopping criterion} \label{sec:convergence}
The DeepSplit algorithm converges to the optimal solution of the convex problem \eqref{eq: verification problem split convexified} under mild conditions. Specifically, when $J$ is closed, proper and convex, and when the sets $\mathcal{S}_k$ (convex outer approximations of the graph of the layers) along with $\mathcal{X}$ are closed and convex, we can resort to the convergence results of ADMM \cite{boyd2011distributed}. Convergence of our algorithm is formally analyzed in Appendix~\ref{app:convergence}.

Following from~\cite{boyd2011distributed}, for the LP relaxation~\eqref{eq: verification problem split convexified} of a feed-forward neural network,  the primal and dual residuals are defined as
\begin{equation}\label{eq:primal_residual}
\begin{aligned} 
r_p &= \sum_{k=0}^{\ell-1} \ \{\|y_{k}^{+}-x_k^{+}\|_2^2 +\|x_{k+1}^{+}-z_{k}^{+}\|_2^2\}, \\
r_{d} &= \rho \sum_{k=1}^{\ell-1} \| (y_k^{+}-y_k)  + (z_{k-1}^+ - z_{k-1}) \|_2^2  + \rho \left(\|y_{0}^{+}-y_{0}\|_2^2 + \| z_{\ell-1}^+ - z_{\ell-1} \|_2^2 \right). \notag
\end{aligned}
\end{equation}
These are the residuals of the optimality conditions for \eqref{eq: verification problem split convexified} and converge to zero as the algorithm proceeds.
A reasonable termination criterion is that the primal and dual residuals must be small, i.e. $r_p \leq \epsilon_p$ and $r_d \leq \epsilon_d$, where $\epsilon_p>0$ and $\epsilon_d>0$ are tolerance levels \cite[Chapter 3]{boyd2011distributed}. These tolerances
can be chosen using an absolute and relative criterion, such as
\begin{equation*}
\begin{aligned}
\epsilon_p \!&=\! \sqrt{p} \ \epsilon_{abs} \!+\! \epsilon_{rel} \max\lbrace(\|x_0\|_2^2 \!+\! 2\sum_{i=1}^{\ell-1} \|x_i\|_2^2 \!+\! \|x_\ell\|_2^2)^{\frac{1}{2}} +(\sum_{i=0}^{\ell-1}(\| y_i\|_2^2 + \|z_i\|_2^2)^{1/2} \rbrace, \notag \\
\epsilon_d \! &=\! \sqrt{n} \ \epsilon_{abs} \! + \! \epsilon_{rel} (\|\lambda_0 \|_2^2 \!+\! \sum_{i=1}^{\ell-1} \| \lambda_i \!+\! \mu_{i-1} \|_2^2 + \|\mu_{\ell-1}\|_2^2)^{\frac{1}{2}}, \notag
\end{aligned}
\end{equation*}
where $p = n_0 + 2 \sum_{i=1}^{\ell-1} n_i + n_\ell$, $n = \sum_{i=0}^\ell n_i$, $\epsilon_{abs}>0$ and $\epsilon_{rel}>0$ are absolute and relative tolerances. Here $n$ is the dimension of $\bx$, the vector of primal variables that are updated in the first step of the algorithm, and $p$ is the total number of consensus constraints. 


\subsection{Residual balancing for convergence acceleration}
A proper selection of the augmentation constant $\rho$ has a dramatic effect on the convergence of the ADMM algorithm. Large values of  $\rho$ enforce consensus more quickly,  yielding smaller primal residuals but larger dual ones. Conversely, smaller values of $\rho$ lead to larger primal and smaller dual residuals. Since ADMM terminates when both the primal and dual residuals are small enough, in practice we prefer to choose the augmentation parameter $\rho$ not too large or too small in order to balance the reduction in the primal and dual residuals. A commonly used heuristic to make this trade-off is residual balancing~\cite{he2000alternating}, in which the penalty parameter varies adaptively based on the following rule:
\begin{align*}
\rho^{+} = \begin{cases} \tau \rho & \text{ if } r_p > \mu r_d \\
\rho/\tau & \text{ if } r_d > \mu r_p 
\\
\rho & \text{otherwise},
\end{cases}
\end{align*}
where $\mu,\tau>1$ are given parameters. In our experiments, we set $\tau = 2, \mu = 10$ and found this rule to be effective in speeding up the practical convergence which is demonstrated numerically in Section~\ref{sec:expr_residual_balancing}.

\section{Connection to Lagrangian-based methods}
\label{sec:lagrangian}
In this section, we consider verification of the feed-forward neural network~\eqref{eq: DNN model} and draw connections to two related methods relying on Lagrangian relaxation. Specifically, we relate our approach to an earlier dual method \cite{dvijotham2018dual} (Section~\ref{sec:dj}), as well as a recent Lagrangian decomposition method  \cite{bunel2020lagrangian} (Section~\ref{sec:bunel}). Overall, these approaches use a similar Lagrangian formulation, but the specific choices in splitting and augmentation of the Lagrangian result in slower theoretical convergence guarantees when solving the convex relaxation to optimality. 
%
%
%

\subsection{Dual method via Lagrangian relaxation of the nonconvex problem}
\label{sec:dj}
Instead of splitting the neural network equations~\eqref{eq: nn verification problem constrained} with auxiliary variables, an alternative strategy is to directly relax~\eqref{eq: nn verification problem constrained} with Lagrangian multipliers \cite{dvijotham2018dual}: 
\begin{align}
 \mathcal{L}(\bx,\blambda) \!=\! J(x_{\ell}) + \sum_{k=0}^{\ell-1} \lambda_k^\top (x_{k+1}\!-\!\phi_k(x_k)) + \mathbb{I}_{\mathcal{X}}(x_0).   
\end{align}
where $\bx = (x_0,\cdots,x_{\ell})$ and $\blambda = (\lambda_0,\cdots,\lambda_{\ell-1})$. This results in the dual problem $g^\star \leftarrow \text{maximize} \ g(\blambda)$, 
%
%
where the dual function is
\begin{align*}
    g(\blambda) &\!=\! \inf_{\underline{x}_{\ell} \leq x_{\ell} \leq \bar{x}_{\ell}} \ \{J(x_{\ell}) + \lambda_{\ell-1}^\top x_{\ell}\} \!+\! \inf_{x_0 \in \mathcal{X}_0} \{-\lambda_0^\top \phi_0(x_0)\}  +\sum_{k=1}^{\ell-1} \inf_{\underline{x}_k \leq x_k \leq \bar{x}_k} \{\lambda_{k-1}^\top x_k \!-\! \lambda_k^\top \phi_k(x_k)\}.
\end{align*}
By weak duality, $g^\star \leq J^\star$. 
The inner minimization problems to compute the dual function $g(\blambda)$ for a given $\blambda$ can often be solved efficiently or even in closed-form \cite{dvijotham2018dual}. The resulting dual problem is unconstrained but non-differentiable; hence it is solved using dual subgradient method \cite{dvijotham2018dual}. However, subgradient methods are known to be very slow with convergence rate $O(1/\sqrt{N})$ where $N$ is the number of updates \cite{nesterov2003introductory}, making it inefficient to find exact solutions to the convex relaxation. On the other hand, this method can be stopped at any time to obtain a valid lower bound. 

\subsection{Lagrangian method via a non-separable splitting}
\label{sec:bunel}
Another related approach is the Lagrangian decomposition method from \cite{bunel2020lagrangian}. 
To decouple the constraints for the convex relaxation of \eqref{eq: nn verification problem constrained}, this approach can be viewed as introducing \emph{one} set of intermediate variables $y_k$ as copies of $x_k$  to obtain
\begin{alignat}{3} \label{eq: nn verification problem split bunel}
J_{\mathrm{relaxed}}^\star \leftarrow &\text{minimize} \quad && J(y_{\ell})  \\ \notag
&\text{subject to } \quad && (y_{k},x_{k+1})\in \mathcal{S}_{\phi_k} \quad &&k=0,\cdots,\ell-1 \\ \notag
& \quad && y_k = x_k \quad && k=0,\cdots,\ell \\ \notag
& \quad && x_0 \in \mathcal{X} \notag
\end{alignat}
This splitting is in the spirit of the splitting introduced in \cite{bunel2020lagrangian,bunel2020branch},\footnote{If we define $\phi_k(x_k) = W_{k+1} \sigma(x_k)+b_{k+1}$, where $W_{k+1},b_{k+1}$ are the parameters of the affine layer and $\sigma$ is a layer of activation functions, this splitting coincides with the one proposed in \cite{bunel2020lagrangian,bunel2020branch}}  and differs from our splitting which uses \emph{two} sets of variables in \eqref{eq: verification problem split}.
By relaxing the consensus constraints $y_k=x_k$, the Lagrangian is
\begin{align} \label{eq: lagrangian of convex relaxation bunel}
    \mathcal{L}(\bx,\by,\bmu) &= J(y_{\ell}) + \sum_{k=0}^{\ell} \mu_k^\top (y_k - x_k) + \sum_{k=0}^{\ell-1} \mathbb{I}_{\mathcal{S}_{\phi_k}}(y_k,x_{k+1})+\mathbb{I}_{\mathcal{X}}(x_0).
\end{align}
Again the Lagrangian is separable and its minimization results in the following dual function
\begin{align*} \label{eq: dual of convex relaxation bunel}
    g(\bmu) &= \inf_{\underline{x}_{\ell} \leq y_{\ell} \leq \bar{x}_{\ell}}  \{J(y_{\ell}) + \mu_{\ell}^\top y_{\ell}\} +\inf_{x_0 \in \mathcal{X}} \{-\mu_0^\top x_0\} + \sum_{k=0}^{\ell-1} \inf_{(y_k,x_{k+1})\in \mathcal{S}_{\phi_k}} \{\mu_k^\top y_k - \mu_{k+1}^\top x_{k+1}\}. 
\end{align*}

Since the dual function is not differentiable, it must be maximized by a subgradient method, which again has an $O(1/\sqrt{N})$ rate. To induce differentiability in the dual function and improve speed, \cite{bunel2020lagrangian} uses augmented Lagrangian. Since only one set of variables was introduced in \eqref{eq: nn verification problem split bunel}, the augmented Lagrangian is no longer separable across the primal variables. 
Therefore, for each update of the dual variable, the augmented Lagrangian must be minimized iteratively. To this end, \cite{bunel2020lagrangian} uses the Frank-Wolfe Algorithm in a block-coordinate as an iterative subroutine. However, this slows down overall convergence and suffers from compounding errors when the sub-problems are not fully solved. When stopping early, the primal minimization must be solved to convergence in order to compute the dual function and produce a valid bound. 

In contrast to the approach described above, in this paper we used a different variable splitting scheme in \eqref{eq: verification problem split} that allows us to \emph{fully} separate layers in a neural network. This subtle difference has a significant impact: we can efficiently minimize the corresponding augmented Lagrangian in \emph{closed form}, without resorting to any iterative subroutine. Specifically, we use the ADMM algorithm, which is known to converge at an  $O(1/N)$ rate \cite{he20121}. In summary, our method enjoys an order of magnitude faster theoretical convergence, is more robust to numerical errors, and has minimal requirements for parameter tuning. {We remark that during the updates of ADMM, the objective value is not necessarily a lower bound on $J^\star$, and hence, one must run the algorithm until convergence to produce such a bound.} To stop early, we can use a similar strategy as the Frank-Wolfe approach from \cite{bunel2020lagrangian} and run the primal iteration to convergence with fixed dual variables in order to compute the dual function, which is a lower  bound on $J^\star$. 

\section{Experiments}
\label{sec:experiment}
The strengths of our method are (a) its ability to exactly solve LP relaxations and (b) do so at scales. To evaluate this, we first demonstrate how solving the LP to optimality leads to tighter certified robustness guarantees in image classification and reinforcement learning tasks (Section~\ref{sec:tightness}) compared with the scalable fast linear bounds methods. We then stress test our method in both speed and scalability against a commercial LP solver (Section~\ref{sec:expr_speed}) and in the large network setting, e.g., solving the LP relaxation for a standard ResNet18 (Section~\ref{sec:expr_scalability}). {In Section~\ref{sec:reachability}, we apply our method on reachability analysis of neural network dynamical systems and compare with the state-of-the-art complete verification method $\alpha,\beta$-CROWN~\cite{wang2021beta}.} Section~\ref{sec:expr_residual_balancing} demonstrates the effectiveness of residual balancing in our numerical examples. 

\paragraph{Setup} In all the experiments, we focus on the setting of \emph{verification-agnostic} networks which are networks trained without promoting verification performances, similar to \cite{dathathri2020enabling}. 
All the networks have been trained adversarially with the projected gradient descent (PGD) attack~\cite{madry2017towards}. 

In all of the test accuracy certification results reported in this paper, we initialize $\rho = 1.0$ and apply residual balancing when running ADMM. In different experiments, the stopping criterion parameters $\epsilon_\text{abs}, \epsilon_\text{rel}$ of ADMM are chosen by trial-and-error to achieve a balance between the accuracy and the runtime of the algorithm. In all these experiments the objective functions are linear~\cite{kolter2017provable}. Full details about the network architecture and network training can be found in Appendix~\ref{app:expr}. 

\subsection{Improved bounds from exact LP solutions}
\label{sec:tightness}
We first demonstrate how solving the LP exactly with our method results in tighter bounds than prior work that solve convex relaxations in a scalable manner. We consider two main settings: certifying the robustness of classifiers for CIFAR10 and deep Q-networks (DQNs) in Atari games. In both experiments, the pre-activation bounds for formulating the LP verification problem are obtained by the fast linear bounds developed in~\cite{kolter2017provable}.

\paragraph{CIFAR10}


We consider a convolutional neural network (CNN) of around 60k hidden units whose convex relaxations cannot be feasibly solved by Gurobi (for LP relaxation) or SDP solvers (for the SDP relaxation). Up until this point, the only solutions for large networks were {linear bounds~\cite{wong2018scaling,xu2020automatic} and Lagrangian-based specialized solvers~\cite{dvijotham2018dual, bunel2020lagrangian}} to the LP relaxation.  



In Table~\ref{tab:cifar10}, we report the certified accuracy of solving the LP exactly with ADMM in comparison to a range of baselines. Verification of the CNN with $\ell_\infty$ perturbation at the input image with different radii $\epsilon$ is conducted on the $10,000$ test images from CIFAR10, and certified test accuracy is reported as the percentage of verified robust test images by each method. We compare with methods that have previously demonstrated the ability to bound networks of this size: fast bounds of the LP (Linear) \cite{wong2018scaling,xu2020automatic}, and interval bounds (IBP) \cite{gowal2018effectiveness}. We additionally compare to a suite of Lagrangian-based baselines, whose effectiveness at this scale was previously unknown. These methods~\footnote{These Lagrangian-based baselines were implemented using the codes at \url{https://github.com/oval-group/decomposition-plnn-bounds}} include supergradient ascent (Adam)~\cite{bunel2020branch}, dual supergradient ascent (Dual Adam)~\cite{dvijotham2018dual} and a variant thereof (Dual Decomp Adam)~\cite{bunel2020branch}, and a proximal method (Prox)~\cite{bunel2020lagrangian}. As mentioned in Section~\ref{sec:bunel}, these baselines require solving an inner optimization problem through the iterations of the algorithms. In the experiments of Table~\ref{tab:cifar10}, the number of iterations of different algorithms are bounded separately such that each Lagrangian-based method has an average runtime of $9$ seconds to finish verifying one example. Our ADMM solver averages $9$ seconds runtime per example in this verification task, which is the same as the average runtime of the Lagrangian-based methods, with the stopping criterion of $\epsilon_{abs} = 10^{-4}, \epsilon_{rel} = 10^{-3}$ and $\rho$ initialized as $1.0$. 

In Table~\ref{tab:cifar10}, we find that solving the LP exactly leads to consistent gains in certified robustness for large networks, with up to {2\%} additional certified robustness over the best-performing alternative. All the methods in Table~\ref{tab:cifar10} are given the same time budget. Indeed, the better theoretical convergence guarantees of ADMM translate to better results in practice: when given a similar budget, the Lagrangian baselines have worse convergence and cannot verify as many examples.  

\begin{remark}
By Table~\ref{tab:cifar10}, our goal is to compare the performances of ADMM and other Lagrangian-based methods in solving the same LP-based neural network verification problem. The certified test accuracy can be further improved if tighter convex relaxations or the BaB techniques are applied. In fact, for the verification problem considered in Table~\ref{tab:cifar10}, the state-of-the-art neural network verification method $\alpha, \beta$-CROWN~\cite{wang2021beta}, which integrates fast linear bounds methods into BaB, is able to achieve certified test accuracies of $65.4\%, 56.0\%, 37.9\%, 19.9\%$ for $\epsilon = 1/255, 1.5/255, 2/255, 2.5/255$, respectively, given the same $9$s time budget per example. However, as will be shown in Section~\ref{sec:reachability}, our method outperforms $\alpha, \beta$-CROWN in reachability analysis of a neural network dynamical system, which highlights the importance of adapting neural network verification tools for tasks of different structures and scales. 
\end{remark}

\begin{table}[tb]
    \centering
    \caption{Certified test accuracy (\%) of PGD-trained models on CIFAR10 through ADMM, the Lagrangian decomposition methods~\cite{dvijotham2018dual,bunel2020branch}, and fast dual/linear \cite{wong2018scaling,xu2020automatic} or interval bounds \cite{gowal2018effectiveness}. All the LP-based methods (ADMM and Lagrangian) are given the same verification time budget of $9$s per example. The fast linear and IBP bounds can be obtained almost instantly in this experiment, and their achievable certified test accuracy is used for reference.}
    \resizebox{\textwidth}{!}{
    \begin{tabular}{lccccccc}
    \toprule
    & Exact & \multicolumn{4}{c}{Lagrangian methods} & \multicolumn{2}{c}{Fast bounds}\\
    \cmidrule(lr){2-2} \cmidrule(lr){3-6} \cmidrule(lr){7-8}
     $\epsilon$ & \hphantom{x} ADMM \hphantom{x} &  \hphantom{x} Adam \hphantom{x} & \hphantom{x} Prox \hphantom{x} & \hphantom{x} Dual Adam \hphantom{x} & \hphantom{x} Dual Decomp Adam \hphantom{x} & \hphantom{x} Linear \hphantom{x} & \hphantom{x} IBP \hphantom{x}\\
     \midrule
     1/255 &  $\mathbf{64.0}$ &  60.5 & 62.4 &   59.8 &  60.3 &  59.8 & 42.8  \\
     1.5/255 & $\mathbf{45.7}$  &  41.2 & 43.5 &   40.5 &   41.1 &  36.8 & 16.8\\
     2/255 & $\mathbf{19.5}$ & 17.3 &  18.2 &  16.9 &  17.1 & 13.2 & 3.6\\
     2.5/255 & $\mathbf{5.5}$  &  4.6 & 4.9 &   4.5 &  4.6 & 3.3 & 0.7 \\
     \bottomrule
\end{tabular} }
    \label{tab:cifar10}
\end{table}

\paragraph{State-robust RL}
We demonstrate our approach on a non-classification benchmark from reinforcement learning: verifying the robustness of deep Q-networks (DQNs) to adversarial state perturbations \cite{zhang2020robust}. Specifically, we verify whether a learned DQN policy outputs stable actions in the discrete space when given perturbed states. Similar to the large network considered in the CIFAR10 setting, this benchmark has only been demonstrably verified with fast but loose linear bounds-based methods\cite{xu2020automatic}. 

We consider three Atari game benchmarks: BankHeist, Roadrunner, and Pong~\footnote{\cite{zhang2020robust} considers one additional RL setting (Freeway). However, the released PGD-trained DQN is completely unverifiable for nearly all epsilons that we considered.} and verify pretrained DQNs which were trained with PGD-adversarial training~\cite{zhang2020robust}. For each benchmark, we verify the robustness of the DQN over $10,000$ randomly sampled frames as our test dataset. 

Similar to the CIFAR10 setting, we observe consistent improvement in certified robustness of the DQN when solving the LP exactly with ADMM across multiple RL settings. We summarize the results using our method and the linear bounds on LP relaxations~\cite{wong2018scaling,xu2020automatic} in Table \ref{tab:rl}.

\begin{table*}[t]
    \centering
   \caption{The percentage of actions from a deep Q-network that are certifiably robust to changes in the state space for three RL tasks: Bankheist, Roadrunner, and Pong. We compare fast linear bounds (Linear)~\cite{wong2018scaling,xu2020automatic} and ADMM.}
   \resizebox{\textwidth}{!}{
    \begin{tabular}{ccc|ccc|ccc}
    \toprule
     \multicolumn{3}{c}{Bankheist} & \multicolumn{3}{c}{Roadrunner} & \multicolumn{3}{c}{Pong} \\
    \midrule
    $\epsilon$ & \hphantom{xxxx} Linear \hphantom{xxxx} & ADMM & $\epsilon$ & \hphantom{xxxx} Linear \hphantom{xxxx} & ADMM & $\epsilon$ & \hphantom{xxxx} Linear \hphantom{xxxx} & ADMM  \\
    \midrule
    0.0016 &  67.0 & \textbf{71.4}  & 0.0012 &  32.6 &   \textbf{36.6} &  0.0004 & 96.1 &  \textbf{97.4}  \\
     0.0020 &  39.7 &  \textbf{49.5} &   0.0016 &  26.3 & \textbf{27.5} &   0.0008&   93.4 &  \textbf{95.6}  \\
     0.0024 & 12.7 & \textbf{25.9} &   0.0020 &   19.6 &  \textbf{22.8} & 0.0012 &  92.1 &\textbf{94.3}  \\
     0.0027 &  1.4 & \textbf{7.3} &0.0024 &   1.1 &   \textbf{3.7} &   0.0016 &82.1 &  \textbf{84.0}  \\
     \bottomrule
    \end{tabular} }
    \label{tab:rl}
\end{table*}


\subsection{Speed}
\label{sec:expr_speed}
%
We compare the solving speeds of our method with state-of-the-art solvers for convex relaxations: a commercial-grade LP solver, Gurobi. Since Gurobi cannot handle large networks, we benchmark the approaches on a fully connected network that Gurobi can handle which is an MNIST network with architecture $784 - 600 - 400 - 200 - 100 - 10$ and ReLU activations (see Appendix~\ref{app:expr} for details). 

To demonstrate the effectiveness of GPU-acceleration in the DeepSplit algorithm, we compare the runtime of DeepSplit and Gurobi in solving LP relaxations that bound the output range of the MNIST network with $\ell_\infty$ perturbations in the input. Specifically, for a given example in the MNIST test data set, we apply DeepSplit/Gurobi layer-by-layer to find the tightest pre-activation bounds under the LP-relaxation. 

With the Gurobi solver, we need to solve $2 \times 600$ LPs sequentially to obtain the lower and upper bounds for the first activation layer, $2 \times 400$ LPs for the second activation layer, and so forth. With DeepSplit, the pre-activation bounds can be computed in batch and allows GPU-acceleration. 

In our experiment, we fix the radius of the $\ell_\infty$ perturbation at the input image as $\epsilon = 0.02$. For the Gurobi solver, we randomly choose $10$ samples from the test data set and compute the pre-activation bounds layer-by-layer. The LP relaxation is formulated in CVXPY and solved by Gurobi v9.1 on an Intel Core i7-6700K CPU, which has 4 cores and 8 threads. For each example, the total solver time of Gurobi is recorded with the average solver time being $275.9$ seconds. For the DeepSplit method, we compute the pre-activation bounds layer-by-layer on $19$ randomly chosen examples. The algorithm applies residual balancing with the initial $\rho = 1.0$ and the stopping criterion is given by $\epsilon_\text{abs} = 10^{-4}, \epsilon_\text{rel}= 10^{-3}$. The total running time of DeepSplit is $717.9$ seconds, with $37.8$ seconds per example on average. With the GPU-acceleration, our method achieves 7x speedup in verifying NN properties compared with the commercial-grade Gurobi solver.

\begin{figure}
	\centering
\begin{subfigure}{0.45 \columnwidth}
    \includegraphics[width = \linewidth]{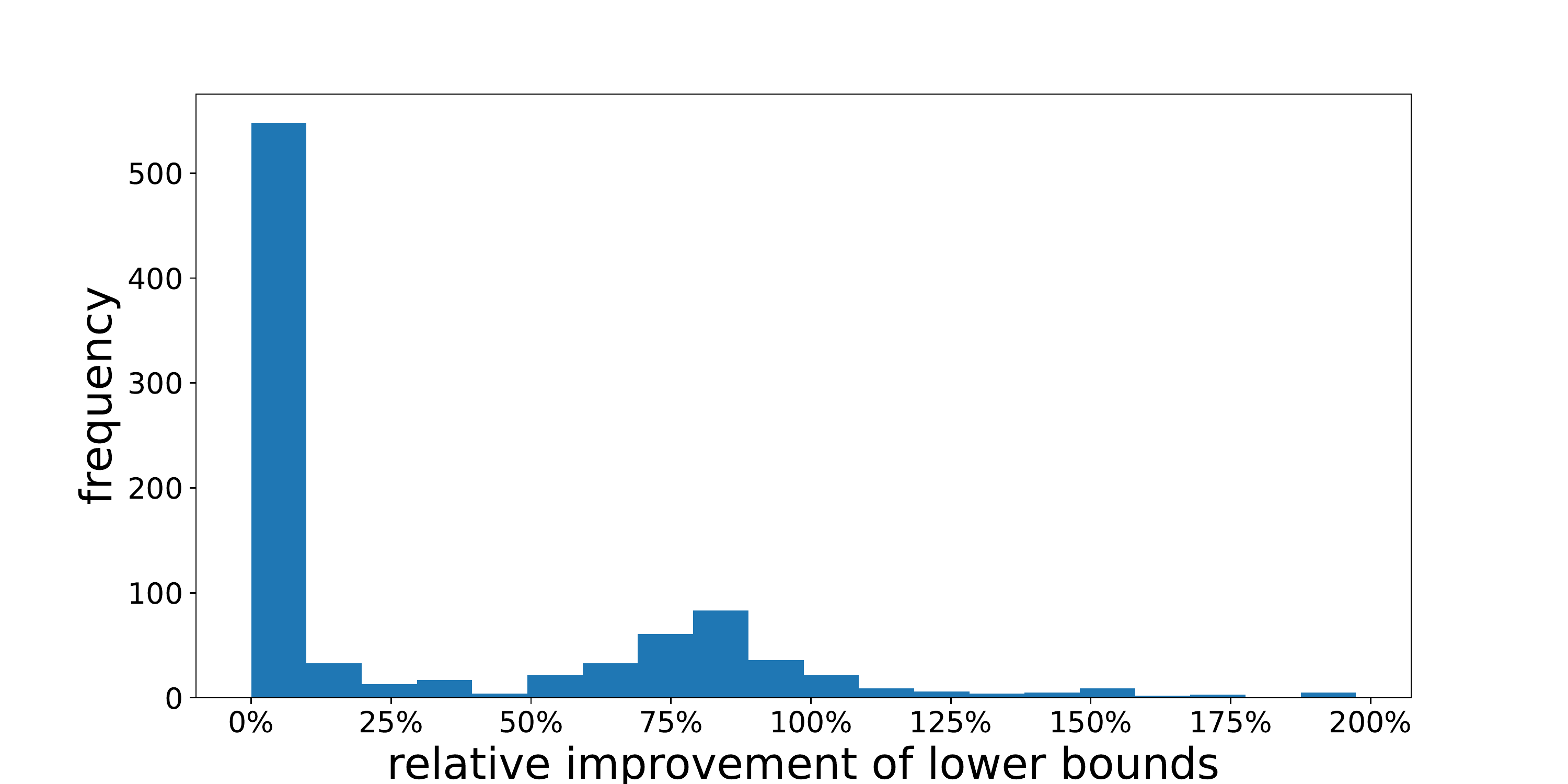}
\end{subfigure}
\hfill
\begin{subfigure}{0.45 \columnwidth}
    \includegraphics[width = \linewidth]{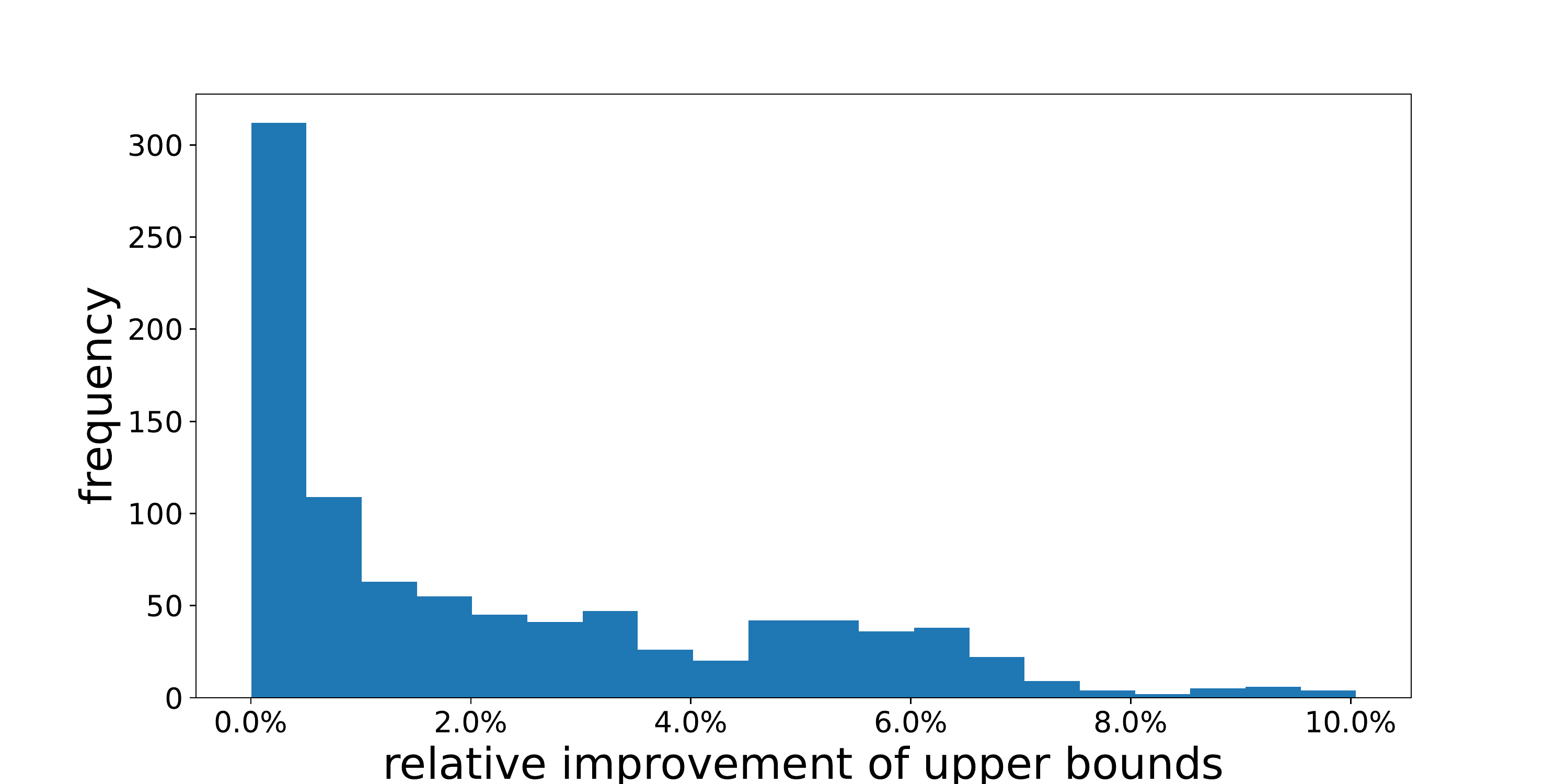}
\end{subfigure}
    \caption{A total of $1000$ ResNet18 output lower and upper bounds are computed from ADMM and LiRPA for comparison in CIFAR10. Histograms of the relative improvement percentage of ADMM over LiRPA are shown for the lower (top) and upper (bottom) bounds, which have an average relative improvement of $31.61\%$ and $2.32\%$, respectively.}
    \label{fig:admm_lirpa}
\end{figure}

\subsection{Scalability}
\label{sec:expr_scalability}
To test the scalability and generality of our approach, we consider solving the LP relaxation for a ResNet18, which up to this point has simply not been possible due to its size. The only applicable method here is LiRPA~\cite{xu2020automatic}---a highly scalable implementation of the linear bounds that works for arbitrary networks but can be quite loose in practice. For this experiment, we measure the improvement in the bound from solving the LP exactly in comparison to LiRPA.

The ResNet18 network is trained on CIFAR10 whose max pooling layer is replaced by a down-sampling convolutional layer for comparison with LiRPA~\cite{xu2020automatic}~\footnote{The max pooling layer has not been considered in the implementation of LiRPA by the submission of this paper. Codes of LiRPA are available at \url{https://github.com/KaidiXu/auto_LiRPA} under the BSD 3-Clause "New" or "Revised" license.} which is capable of computing provable linear bounds for the outputs of general neural networks and is the only method available so far that can handle ResNet18. The ResNet18 is adversarially trained using the fast adversarial training code from \cite{wong2020fast}. 

In our experiments, for the first $100$ test examples in CIFAR10, we use LiRPA to compute the preactivation bounds for each ReLU layer in ResNet18 and then apply ADMM to compute the lower and upper bounds of ResNet18 outputs (there are $10$ outputs corresponding to the $10$ classes of the dataset). The ADMM is run with stopping criterion $\epsilon_{abs} = 10^{-5}, \epsilon_{rel} = 10^{-4}$. With $\ell_\infty$ ball input perturbation of radius $\epsilon  = 1/255$, we find that exact LP solutions with our ADMM solver can produce substantial improvements in the bound at ResNet18 scales, as shown in Figure~\ref{fig:admm_lirpa}. For a substantial number of examples, we find that ADMM can find significantly tighter bounds (especially for lower bounds). 

{
\subsection{Reachability analysis of dynamical systems}
\label{sec:reachability}
We consider over-approximating the reachable sets of a discrete-time neural network dynamical system $x(t+1) = f_{NN}(x(t))$ over a finite horizon where $x(t) \in \mathbb{R}^{n_x}$ denotes the state at time $t=0,1,\cdots$, $f_{NN}$ is a feed-forward neural network, and $n_x$ is the dimension of the system. Specifically, we consider a cart-pole system with $4$ states under nonlinear model predictive control~\cite{lucia2017rapid}, and train a $4-100-100-4$ neural network $f_{NN}(x)$ with ReLU activations to approximate the closed-loop dynamics with sampling time $0.05 s$. The $4$ states $x = [x_1 \ x_2 \ x_3 \ x_4]^\top$ of the cart-pole system represent the position and velocity of the cart, and the angle and angular speed of the pendulum, respectively. Given an initial set $\mathcal{X}_0 \subset \mathbb{R}^4$ such that $x(0) \in \mathcal{X}_0$, we want to over-approximate the reachable set of $x(t) = f_{NN}^{(t)}(x(0))$ where $f_{NN}^{(t)}$ denotes the $t$-th order composition of $f_{NN}$ and is a feed-forward neural network itself.  

Over-approximating $x(t)$ can be formulated as a neural network verification problem~\eqref{eq: maximization problem} where $f = f_{NN}^{(t)}$ is given by the sequential concatenation of $t$ copies of $f_{NN}$ and the input set $\mathcal{X}$ is chosen as the initial set $\mathcal{X}_0$. A box approximation of $x(t)$ can be obtained by minimizing/maximizing the $i$-th output of the neural network $f^{(t)}_{NN}$ for $1 \leq i \leq n_x = 4$.

With a randomly chosen initial set $x(0) \in \mathcal{X}_0 = \{x \mid x_1 \in [-1.8 \ -1.6] \ \textrm{m}, x_2 \in [0.5 \ 0.7] \ \textrm{m/s}, x_3 \in [0.035 \ 0.105]\ \textrm{rad}, x_4 \in [0.2 \ 0.4]\ \textrm{rad/s}\}$, we consider the horizon $t = 20$ and compute a box over-approximation of the reachable set of $x(20)$ through both DeepSplit and the state-of-the-art BaB-based verification method $\alpha, \beta$-CROWN~\cite{wang2021beta} \footnote{Codes of $\alpha,\beta$-CROWN are available at \url{https://github.com/huanzhang12/alpha-beta-CROWN} under the BSD 3-Clause "New" or "Revised" license.}. From $1000$ simulated trajectories with uniformly sampled initial states from $\mathcal{X}_0$, the emperically estimated lower and upper bounds for $x(20)$ are given by $[-1.74 \ 0.14 \ -0.10 \ 0.022]^\top \leq x(20) \leq [-0.91 \ 0.92 \ -0.036 \ 0.11]^\top$. With the time budget $1470$s per bound, the box over-approximations of $x(20)$ given by DeepSplit and $\alpha,\beta$-CROWN are given as follows:
\begin{equation*}
\begin{aligned}
     \textrm{DeepSplit:} &
    \begin{bmatrix}
    -4.19 \\ -4.39 \\ -2.14 \\ -3.88
    \end{bmatrix} \leq x(20) \leq 
    \begin{bmatrix}
    3.00 \\ 4.81 \\ 1.99 \\ 3.91
    \end{bmatrix}, \\
    \alpha, \beta \textrm{-CROWN}: &
        \begin{bmatrix}
    -38.38 \\ -46.52 \\ -23.85 \\ -55.97
    \end{bmatrix} \leq x(20) \leq 
    \begin{bmatrix}
    37.08 \\ 48.58 \\ 22.86 \\ 48.21
    \end{bmatrix}.
\end{aligned}
\end{equation*}
We observe that DeepSplit gives a reasonable over-approximation of the reachable set of $x(20)$, while the bounds obtained by $\alpha,\beta$-CROWN in this task are an order of magnitude more conservative. Such comparison result holds for other randomly chosen initial sets $\mathcal{X}_0$ too. The details of the experimental setup is shown in Appendix~\ref{app:expr}.

}

\subsection{Effects of residual balancing}
\label{sec:expr_residual_balancing}
We demonstrate the effects of residual balancing on the convergence of ADMM through the MNIST network (see Appendix~\ref{app:expr} for details). We conduct our experiment on the $1938$-th example which is randomly chosen from the MNIST test dataset. For this example, the MNIST network predicts its class (number $4$) correctly. We add an $\ell_\infty$ perturbation of radius $\epsilon = 0.02$ to the input image and verify if the network outputs are robust with respect to class number $3$. {This corresponds to setting $i^\star = 4,i = 3$ and $\mathcal{X} = \{x \mid \lVert x - x^* \rVert_\infty \leq 0.02 \}$ in problem~\eqref{eq:class_verify} where $x^*$ denotes the chosen test image.}

\begin{figure*}[tb]
	\centering
	\begin{subfigure}{0.32\textwidth}
		\centering
		\includegraphics[width=\linewidth]{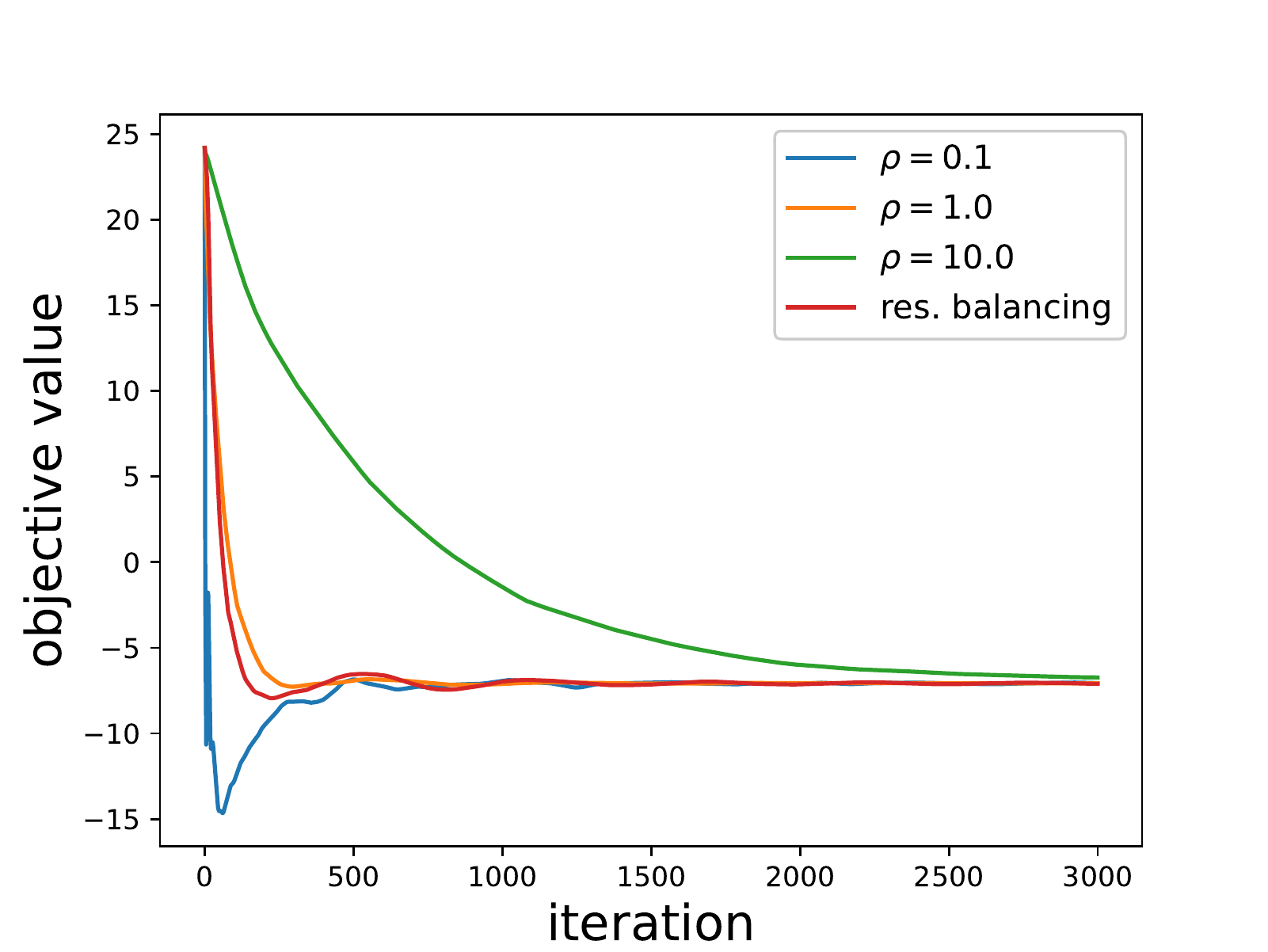}
	\end{subfigure}
	\hfill
	\begin{subfigure}{0.32\textwidth}
		\centering
		\includegraphics[width=\linewidth]{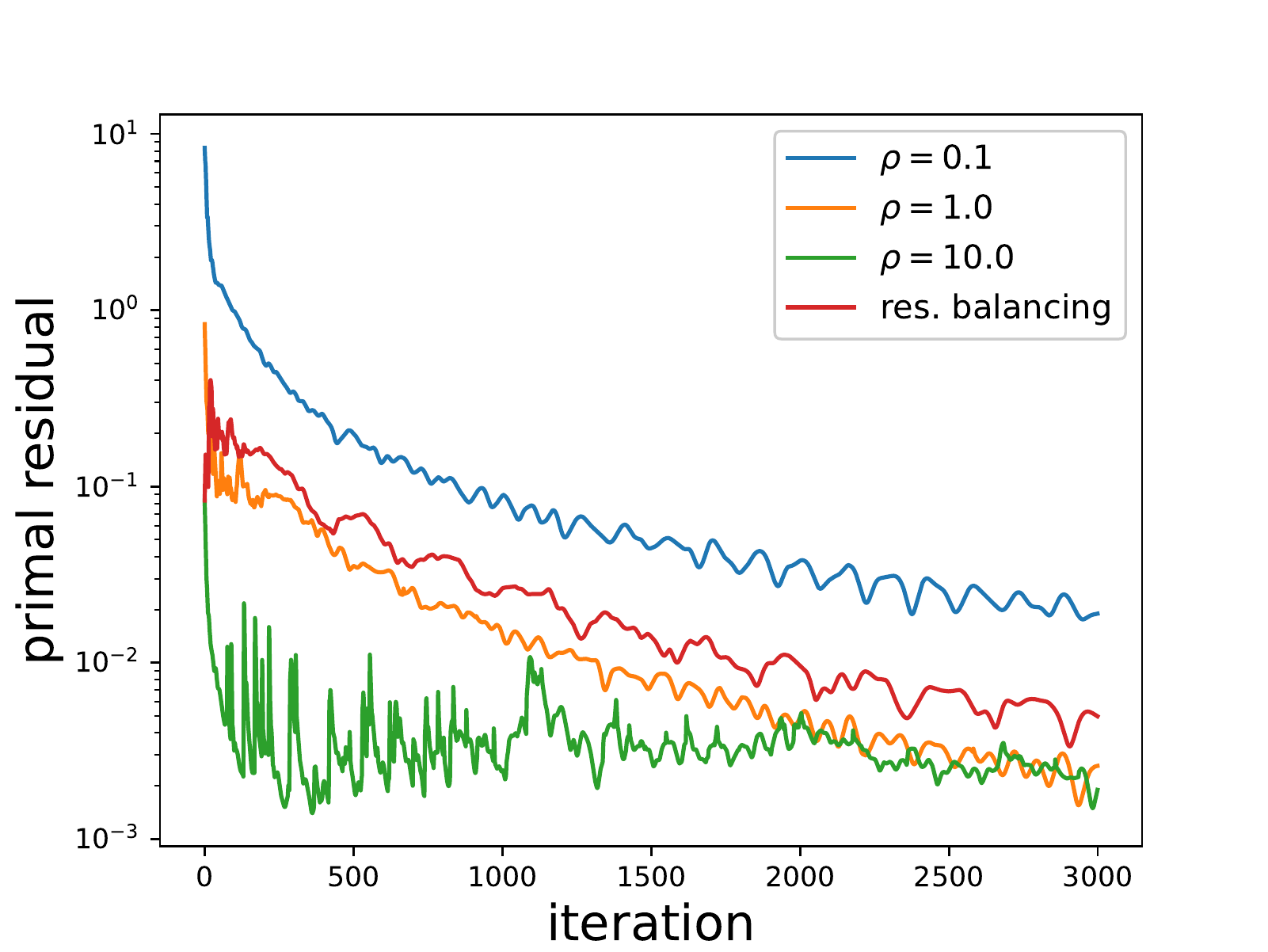}
	\end{subfigure}
	\hfill
	\begin{subfigure}{0.32 \textwidth}
		\centering
		\includegraphics[width = \linewidth]{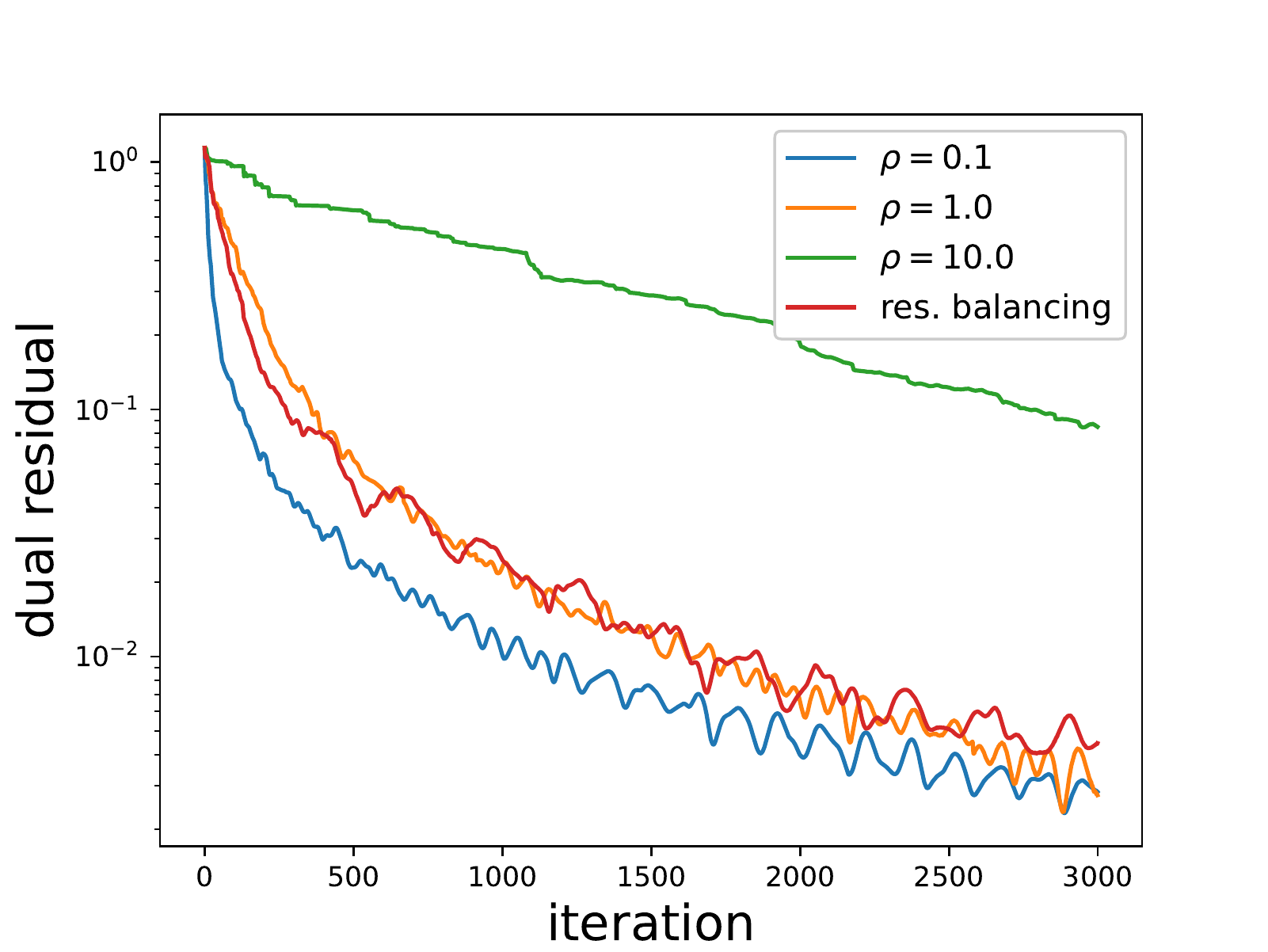}
	\end{subfigure}
	\caption{The objective values (left), primal residuals (middle), and dual residuals (right) of ADMM under different augmentation parameters $\rho$ on the fully connected MNIST  network described in Appendix~\ref{app:expr}. }
	\label{fig:mnist_fc4}
\end{figure*}

The maximum number of iterations is restricted to $3000$. The objective values, primal and dual residuals of ADMM for the network under different fixed augmentation parameters $\rho$ are plotted in Figure~\ref{fig:mnist_fc4}. The residual balancing in this experiment is applied with $\tau = 2$, $\mu = 10$, and $\rho$ initialized as $10.0$. 

The effects of $\rho$ on the convergence rates of the primal and dual residuals are illustrated empirically in Figure~\ref{fig:mnist_fc4}. Despite initialized at a large value $10.0$, residual balancing is able to adapt the value of $\rho$ and achieves significant improvement in convergence rate compared with the case of constant $\rho = 10.0$. As observed in our other experiments, with residual balancing, ADMM becomes insensitive to the initialization of $\rho$ and usually achieves a good convergence rate. 



\section{Conclusion}
In this paper, we proposed DeepSplit, a scalable and modular operator splitting technique for solving convex relaxation-based verification problems for neural networks. 
The method can exactly solve large-scale LP relaxations with GPU acceleration with favorable convergence rates. Our approach leads to tighter bounds across a range of classification and reinforcement learning benchmarks, and can scale to a standard ResNet18. 
%
%
%
We leave as future work a further investigation of variations of ADMM that can improve convergence rates in deep learning-sized problem instances, as well as extensions beyond the LP setting. {Furthermore, it would be interesting to extend the proposed method to verification of recurrent neural neworks (RNNs) such as vanilla RNNs, LSTMs\footnote{Long Short-Term Memory.}, and GRUs\footnote{Gated Recurrent Unit.} \cite{ko2019popqorn,ryou2021scalable,mohammadinejad2021diffrnn}.}


\appendix
\section{Appendix}
\subsection{Convergence analysis of DeepSplit}
\label{app:convergence}
By defining $\bx_1 = \bx, \ \bx_2 = (\by,\bz)$ (the primal variables) and $\bnu = (\blambda,\bmu)$ (the scaled dual variables), we can write the convex relaxation in \eqref{eq: verification problem split convexified} as
\begin{alignat}{2} \label{eq: verification problem split convexified compact}
    &\mathrm{minimize} \ &&f_1(\bx_1) + f_2(\bx_2) \\
    &\text{subject to} \ && A_1 \bx_1+ A_2\bx_2 = 0 \notag
\end{alignat}
with the corresponding Augmented Lagrangian
\begin{align}\label{eq: augmented lagrangian compact}
    \mathcal{L}(\bx_1,\bx_2,\bnu) &= f_1(\bx_1) + f_2(\bx_2) + \frac{\rho}{2}(\|A_1 \bx_1+ A_2\bx_2 +\bnu\|_2^2-\|\bnu\|_2^2) \notag
\end{align}
where $f_1 \colon \mathbb{R}^{n} \to \mathbb{R} \cup \{+\infty\}$ and $f_2 \colon \mathbb{R}^{2n-2n_{\ell}} \to \mathbb{R} \cup \{+\infty\}$ are extended real-valued functions defined as $f_1(\bx_1) :=  J(x_{\ell})+\mathbb{I}_{\mathcal{X}}(x_0)$, $f_2(\bx_2):= \sum_{k=0}^{\ell-1} \mathbb{I}_{\mathcal{S}_{\phi_k}}(y_{k}, z_{k})$. Moreover, $A_1\bx_1+A_2\bx_2=0$ represents the set of equality constraints $y_k=x_k$ and $x_{k+1}=z_{k}$ for $k=0,\cdots,\ell$. The dual function is given by
\begin{align}
    g(\bnu) = \inf_{\bx_1,\bx_2} \mathcal{L}(\bx_1,\bx_2,\bnu).
\end{align}
By Danskin’s theorem \cite{bertsekas1997nonlinear}, the sub-differential of the dual function is given by
\begin{align}
    \partial g(\bnu) = \{A_1 \bar{\bx}_1+ A_2 \bar{\bx}_2 \colon (\bar{\bx}_1,\bar{\bx}_2) \in \arg\min_{\bx_1,\bx_2} \mathcal{L}(\bx_1,\bx_2,\bnu) \}.
\end{align}
Here $(\bar{\bx}_1,\bar{\bx}_2)$ is a minimizer of the Lagrangian (not necessarily unique), which satisfies the optimality conditions
\begin{align} \label{eq: Lagrangian minimizer optimality condition 1}
    0 &\in \partial f_1(\bar{\bx}_1) + \rho A
    _1^\top (A_1\bar{\bx}_1+A_2 \bar{\bx}_2 + \bnu), \\
     0 &\in \partial f_2(\bar{\bx}_2) + \rho A
    _2^\top (A_1\bar{\bx}_1+A_2 \bar{\bx}_2 + \bnu). \notag
\end{align}
We want to show that the sub-differential is a singleton, i.e., $g$ is continuously differentiable. Suppose $(\bar{\bx}_1, \bar{\bx}_2) \in \arg\min_{\bxi} \mathcal{L}(\bxi,\bnu)$ and $(\bar{\bw}_1, \bar{\bw}_2) \in \arg\min_{\bxi} \mathcal{L}(\bxi,\bnu)$ are two distinct minimizers of the Lagrangian, hence satisfying
\begin{align}  \label{eq: Lagrangian minimizer optimality condition 2}
    0 &\in \partial f_1(\bar{\bw}_1) + \rho A
    _1^\top (A_1\bar{\bw}_1+A_2 \bar{\bw}_2 + \bnu), \\
     0 &\in \partial f_2(\bar{\bw}_2) + \rho A
    _2^\top (A_1\bar{\bw}_1+A_2 \bar{\bw}_2 + \bnu). \notag
\end{align}
By monotonicity of the sub-differentials, we can write
{
\begin{align} \label{eq: subdiff monotone}
    &(T_{f_1}(\bar{\bx}_1)-T_{f_1}(\bar{\bw}_1))^\top (\bar{\bx}_1-\bar{\bw}_1) \geq 0, \\
    &(T_{f_2}(\bar{\bx}_2)-T_{f_2}(\bar{\bw}_2))^\top (\bar{\bx}_2-\bar{\bw}_2) \geq 0. \notag
\end{align}
where $T_{f}(\bar{\bx}) \in \partial f(\bar{\bx})$ denotes a subgradient.} By substituting \eqref{eq: Lagrangian minimizer optimality condition 1} and \eqref{eq: Lagrangian minimizer optimality condition 2} in \eqref{eq: subdiff monotone}, we obtain
\begin{align*}
-\rho A_1^\top ((A_1\bar{\bx}_1+A_2 \bar{\bx}_2 + \bnu) -(A_1\bar{\bw}_1+A_2 \bar{\bw}_2 + \bnu))^\top (\bar{\bx}_1-\bar{\bw}_1) \geq 0,\\
-\rho A_2^\top ((A_1\bar{\bx}_1+A_2 \bar{\bx}_2 + \bnu)-(A_1\bar{\bw}_1+A_2 \bar{\bw}_2 + \bnu))^\top (\bar{\bx}_2-\bar{\bw}_2) \geq 0.
\end{align*}

By adding the preceding inequalities, we obtain
\begin{align}
    -\rho \|A_1\bar{\bx}_1+A_2\bar{\bx}_2-(A_1\bar{\bw}_1+A_2\bar{\bw}_2)\|_2^2 \geq 0.
\end{align}
When $\rho>0$, this implies that $A_1\bar{\bx}_1+A_2\bar{\bx}_2=A_1\bar{\bw}_1+A_2\bar{\bw}_2$ and hence, the sub-differential $\partial g$ is a singleton. 

\noindent \textbf{Convergence. } When $\mathcal{X}$ is a closed nonempty convex set, $\mathbb{I}_{\mathcal{X}}(x_0)$ is a convex closed proper (CCP) function. Assuming that $J$ is also CCP, then we can conclude that $f_1$ is CCP. Furthermore, since the sets $\mathcal{S}_{\phi_k}$ are nonempty convex sets, we can conclude that $f_2$ is CCP. Under these assumptions, the augmented Lagrangian has a minimizer (not necessarily unique) for each value of the dual variables. Finally, under the assumption that the Augmented Lagrangian has a saddle point (which produces a solution to \eqref{eq: verification problem split convexified compact}), the ADMM algorithm we have primal convergence $\|r_p\|_2 \to 0$ (see \eqref{eq:primal_residual}), dual residual convergence $\|r_d\|_2 \to 0$, as well as objective convergence $J(x_{\ell}) \to J^\star$ \cite{boyd2011distributed}.

We remark that the convergence guarantees of ADMM holds even if $f_1$ and $f_2$ assume the value $+\infty$. This is the case for indicator functions resulting in projections in the first two updates of Algorithm \ref{DeepSplit Algorithm}. 

\subsection{Projection onto convolutional layers}  
\label{app:convolutions}
Although a convolution is  a linear operator, it is impractical to directly form the inverse matrix for the projection step of the DeepSplit algorithm~\eqref{eq:affine_proj}. Instead, we represent a typical convolutional layer $f_\text{conv}$ with stride, padding and bias as
\begin{equation} \label{eq:conv_decomposition}
   \begin{aligned}
   f_\text{conv} &= f_\text{bias}\circ f_\text{ds} \circ f_\text{crop} \circ f_\text{circ} \circ  f_\text{pad}
   = f_\text{post}\circ f_\text{circ} \circ f_\text{pad},
   \end{aligned}
   \end{equation}
where $f_\text{pad}$ is a padding step, $f_\text{circ}$ is a circular convolution step, $f_\text{crop}$ is a cropping step, $f_\text{ds}$ is a downsampling step to handle stride greater than one, and $f_\text{bias}$ is a step that adds the bias. In other words, a convolutional layer can be decomposed into five sequential layers in our neural network representation~\eqref{eq: DNN model}. In practice, we combine the last three steps into one post-processing layer $f_\text{post}$ to reduce the number of concensus constraints in the DeepSplit algorithm. The projection steps for all of these operators are presented next. An efficient FFT implementation of the projection step for the affine layer $f_\text{circ}$ is given in Appendix~\ref{app:FFT}.

\paragraph{Padding}
The padding layer $f_\text{pad}$ takes an image as input and adds padding to it. Denote the input image by $y_k$ and the padded image by $z_k$. We can decompose the output $z_k$ into two vectors, $z^0_k$ which is a copy of the input $y_k$, and $z^1_k$ which represents the padded zeros on the edges of image. Equivalently, the padding layer $z_k = \phi_k(y_k)$ can be written in an affine form
\begin{equation*}
    z_k = \begin{bmatrix} z^0_k \\ z^1_k \end{bmatrix} =  \begin{bmatrix} I \\ 0 \end{bmatrix} y_k = W_k y_k,
\end{equation*}
for which the projection operator reduces to the affine case. 


\paragraph{Cropping}
The cropping layer $f_\text{crop}$ crops the output of the circular convolution $f_\text{circ}$ to the original size of the input image before padding. Denote $y_k$ the input image and $z_k$ the output image of the cropping layer. By decomposing the input image $y_k$ into the uncropped pixels $y^0_k$ and the cropped pixels $y^1_k$, the cropping layer $z_k = \phi_k(y_k)$ has an affine formulation
\begin{align*}
    z_k = y^0_k = \begin{bmatrix} I  & 0 \end{bmatrix} \begin{bmatrix} y^0_k \\ y^1_k \end{bmatrix} = W_k y_k
\end{align*}
whose projection operator is given in Section~\ref{sec:yz_update}.

\paragraph{Down-sampling and bias}
If the typical convolutional layer $f_\text{conv}$ has stride greater than one, a down-sampling layer is added in the DeepSplit algorithm, which essentially has the same affine form as the cropping layer with different values of $y^0_k$ and $y^1_k$. Therefore, the projection operator for the down-sampling layer reduces to the affine case as well.

The bias layer in the DeepSplit algorithm handles the case when the convolutional layer $f_\text{conv}$ has a bias $b_k$ and is implemented by $z_k = \phi_k(y_k) = y_k + b_k$. This is an affine expression and its projection operator is given in Section~\ref{sec:yz_update}.

\paragraph{Convolutional post-processing layer}
We combine the cropping, down-sampling and bias layers into one post-processing layer, i.e., $f_\text{post} = f_\text{bias} \circ f_\text{ds} \circ f_\text{crop}$, as shown in~\eqref{eq:conv_decomposition}. This reduces the total number of concensus constraints in the DeepSplit algorithm. Since all the three layers are in fact affine, the post-processing layer is also affine and its projection operator can be obtained correspondingly.

\subsection{FFT implementation for circular convolutions}
\label{app:FFT}
In order to efficiently implement projection onto the convolutional layer~\eqref{eq:conv_decomposition}, recall that we can decompose a convolution into the following three steps: 
\begin{align}
    f_\text{conv} = f_\text{post}\circ f_\text{circ} \circ f_\text{pad}.
\end{align}
We now discuss in detail how to efficiently perform the $(y,z)$ update for multi-channel, circular convolutions $f_\text{circ}$ using Fourier transforms.  We begin with the single-channel setting, and then extend our procedure to the multi-channel setting. 

\paragraph{Single-channel circular convolutions}
Let $U$ represent the discrete Fourier transform (DFT) as a linear operator, and let $W$ be the weight matrix for the circular convolution $f_\text{circ}(x) = W * x$. Then, using matrix notation, the convolution theorem states that 
\begin{align}
f_\text{circ}(x) = W * x = U^*(UW\cdot Ux) = U^* D Ux
\end{align}
where $D = \textrm{diag}(UW)$ is a diagonal matrix containing the Fourier transform of $W$ and $U^*$ is the conjugate transpose of $U$.
Then, we can represent the inverse operator from \eqref{eq:affine_proj} as 
\begin{align}
(I + f_\text{circ}^\top f_\text{circ})^{-1} =
U^*(I + DD)^{-1}U.
\end{align}
Since $(I+DD)$ is a diagonal matrix, its inverse can be computed by simply inverting the diagonal elements, and requires storage space no larger than the original kernel matrix. Thus, multiplication by the inverse matrix for a circular convolution reduces to two DFTs and  an element-wise product. For an input of size $n\times n$, this step has an overall complexity of $O(n^2 \log n)$ when using fast Fourier transforms. 

\paragraph{Multi-channel circular convolutions}
We now extend the operation for single-channel circular convolutions to multi-channel, which is typically used in convolutional layers found in deep vision classifiers. Specifically, for a circular convolution with $n$ input channels and $m$ output channels, we have 
\begin{align}
f_\text{circ}(x)_j = \sum_{i=1}^{n} W_{ij} * x_i
\end{align}
where $f_\text{circ}(x)_j$ is the $j$th output channel output of the circular convolution, $W_{ij}$ is the kernel of the $i$th input channel for the $j$th output channel, and $x_i$ is the $i$th channel of the input $x$. The convolutional theorem again tells us that 
\begin{align}
f_\text{circ}(x)_j = \sum_{i=1}^{n} U^* D_{ij} Ux_i
\end{align}
where $D_{ij} = \textrm{diag}(UW_{ij})$. This can be re-written more compactly using matrices as 
\begin{align}
f_\text{circ}(x) = \bar{U}^* \bar{D} \bar{U}\bar{x}
\end{align}
where 
\begin{itemize}
    \item $\bar{U} = \left[\begin{array}{ccc}
     U & \cdots & 0  \\
     \vdots & \ddots & \vdots \\
     0 & \cdots & U
\end{array}\right]$ is a block diagonal matrix with $n$ copies of $U$ along the diagonal
    \item $\bar{U}^* = \left[\begin{array}{ccc}
     U^* & \cdots & 0  \\
     \vdots & \ddots & \vdots \\
     0 & \cdots & U^*
\end{array}\right]$ is a block diagonal matrix with $m$ copies of $U$ along the diagonal
    \item $\bar{D} = \left[\begin{array}{ccc}
     D_{11} & \cdots & D_{n1}  \\
     \vdots & \ddots & \vdots \\
     D_{1m} & \cdots & D_{nm}
\end{array}\right]$ is a block matrix with diagonal blocks where the $ij$th block is $D_{ij}$
    \item $\bar{x} = \left[\begin{array}{c}
     x_1 \\
     \vdots \\
     x_n
\end{array}\right]$ is a vertical stacking of all the input channels. 
\end{itemize}
Then, we can represent the inverse operator from \eqref{eq:affine_proj} as 
\begin{align}
(I + f_\text{circ}^\top f_\text{circ})^{-1} =
\bar{U}^*(I + \bar{D}\bar{D})^{-1}\bar{U}
\end{align}
where $I + \bar{D}\bar{D}$ is a block matrix, where each block is a diagonal matrix. The inverse can then be calculated by the inverting sub-matrices formed from sub-indexing the diagonal components. Specifically, let $\bar{D}_{j::p}$ be a slice of $\bar{D}$ containing elements spaced $m$ elements apart in both column and row directions, starting with the $j$th item. For example, $\bar{D}_{0::p}$ is the matrix obtained by taking the top-left most element along the diagonal of every block. Then, for $j = 1 \dots m$, we have 
\begin{align}
    (I + \bar{D}\bar{D})^{-1}_{j::p} = \left((I + \bar{D}\bar{D})_{j::p}\right)^{-1}.
\end{align}
Thus, calculating this matrix amounts to inverting a batch of $p$ matrices of size $m\times m$. For typical convolutional networks, $m$ is typically well below $1,000$, and so this can be calculated quickly. Further note that this only needs to be calculated once as a pre-computation step, and can be reused across different inputs and objectives for the network. 

\paragraph{Memory and runtime requirements}
In practice, we do not store the fully-expanded block diagonal matrices; instead, we omit the zero entries and directly store the the diagonal entries themselves. Consequently, for an input of size $p$, the diagonal matrices require storage of size $O(mnp)$, and the inverse matrix requires storage of size $O(m^2p)$. Since the discrete Fourier transform can be done in $O(p\log p)$ time with fast Fourier transforms, the overall runtime of the precomputation step to form the matrix inverse is the cost of the initial DFT and the batch matrix inverse, or $O(nmp\log p + m^3p)$. Finally, the runtime of the projection step is $O((n+m)p\log p + n^2mp)$, which is the respective costs of the DFT transformations $\bar U$ and $\bar U^*$, as well as the multiplication by $\bar D$. Since the number of channels in a deep network are typically much smaller than the size of the input to a convolution (i.e. $n < p$ and $m < p$), the costs of doing the cyclic convolution with Fourier transforms are in line with typical deep learning architectures.

\subsection{Experimental details}
\label{app:expr}
\paragraph{CIFAR10}
For CIFAR10, we use the large convolutional architectures from~\cite{wong2018scaling}, which consists of four convolutional layers with $32-32-64-64$ channels, with strides $1-2-1-2$, kernel sizes $3-4-3-4$, and padding $1-1-1-1$. This is followed by three linear layers of size $512-512-10$. This is significantly larger than the CIFAR10 architectures considered by \cite{dathathri2020enabling}, and has sufficient capacity to reach $43\%$ adversarial accuracy against an $\ell_\infty$ PGD adversary at $\epsilon=8/255$. 

The model is trained against a PGD adversary with 7 steps of size $\alpha=2/255$ at a maximum radius of $\epsilon=8.8/255$, with batch size 128 for 200 epochs. We used the SGD optimizer with a cyclic learning rate (maximum learning rate of 0.23), momentum 0.9, and weight decay $0.0005$. The model achieves a clean test accuracy of 71.8\%. 

\paragraph{State-robust RL}
We use the pretrained, adversarially trained, DQNs released by \cite{zhang2020robust}~\footnote{Available at \url{https://github.com/chenhongge/SA_DQN}.}. These models were trained to be robust at $\epsilon=1/255$ with a PGD adversary for the Atari games Pong, Roadrunner, Freeway, and BankHeist. Each input to the DQN is of size $1 \times 84 \times 84$, which is more than double the size of CIFAR10. The DQN architectures are convolutional networks, with three convolutional layers with $32-64-64$ channels, with kernel sizes $8-4-3$, strides $4-2-1$, and no padding. This is followed by two linear layers of size $512-K$, where $K$ is the number of discrete actions available in each game. 

\paragraph{MNIST}
For MNIST, we consider a fully connected network with layer sizes $784-600-400-200-100-10$ and ReLU activations. It is more than triple the size of that considered by \cite{dathathri2020enabling} with one additional layer. It is, however, still small enough such that Gurobi is able to solve the LP relaxation, and allows us to compare our running time against Gurobi. We train the network with an $\ell_\infty$ PGD adversary at radius $\epsilon=0.1$, using 7 steps of size $\alpha=0.02$, with batch size 100 for 100 epochs. We use the Adam optimizer with a cyclic learning rate (maximum learning rate of 0.005), and both models achieve a clean accuracy of 99\%.

\paragraph{ResNet18}
To highlight the scalability of our approach, we consider a ResNet18 network trained on CIFAR10 whose max pooling layer is replaced by a down-sampling convolutional layer for comparison with LiRPA~\cite{xu2020automatic}~\footnote{Codes available at \url{https://github.com/KaidiXu/auto_LiRPA}.}, which is capable of computing provable linear bounds for the outputs of general neural networks and is the only method available so far that can handle ResNet18. The ResNet18 is adversarially trained using the fast adversarial training code from \cite{wong2020fast}. 

{
\paragraph{Reachability analysis example}
Following the notation from Section~\ref{sec:reachability}, the neural network dynamics $f_{NN}$ has the architecture $4-100-100-4$ with ReLU activations and is trained over $10000$ randomly collected state transition samples of the closed-loop dynamics over a bounded domain in the state space. With DeepSplit, we solve the LP-based verification problem~\eqref{eq: verification problem split convexified} on truncated neural networks to find all the pre-activation bounds in $f_{NN}^{(20)}$, while $\alpha, \beta$-CROWN is directly run on $f_{NN}^{(20)}$ since it searches pre-activation bounds automatically. Note that the same time budget of $1470$s per bound is assigned for both methods. The ADMM is run with stopping criterion $\epsilon_{abs} = 10^{-5}, \epsilon_{rel} = 10^{-4}$, while $\alpha, \beta$-CROWN is run using the default configuration parameters~\footnote{Available at \url{https://github.com/huanzhang12/alpha-beta-CROWN}.}.
}



\bibliographystyle{ieeetr}
\bibliography{refs}

\end{document}